\documentclass[10pt,twocolumn,letterpaper]{article}

\usepackage[pagenumbers]{cvpr} 

\usepackage[dvipsnames]{xcolor}

\usepackage{times}
\usepackage{epsfig}
\usepackage{graphicx,graphics}
\usepackage{amsmath}
\usepackage{amssymb}
\usepackage{booktabs}
\usepackage{caption}
\usepackage{algorithm}
\usepackage{algorithmic}
\usepackage{framed}
\usepackage{arydshln}
\usepackage{multirow}
\usepackage{multicol}
\usepackage{microtype}
\usepackage{booktabs,bigstrut}
\usepackage{enumitem}
\usepackage{amsthm}
\usepackage{wrapfig}

\usepackage[accsupp]{axessibility}

\usepackage{color, xcolor, colortbl}
\definecolor{cvprblue}{rgb}{0.21,0.49,0.74}
\usepackage[pagebackref,breaklinks,colorlinks,citecolor=cvprblue]{hyperref}

\newcommand{\bfsection}[1]{\vspace*{0.00cm}\noindent\textbf{#1.}}

\newtheorem{definition}{Definition}
\newtheorem{theorem}{Theorem}
\newtheorem{remark}{Remark}
\usepackage[capitalize]{cleveref}
\crefname{section}{Sec.}{Secs.}
\Crefname{section}{Section}{Sections}
\Crefname{table}{Table}{Tables}
\crefname{table}{Tab.}{Tabs.}

\graphicspath{{Figure/}}
\definecolor{cvprblue}{rgb}{0.21,0.49,0.74}

\usepackage[pagebackref,breaklinks,colorlinks,citecolor=cvprblue]{hyperref}

\def\paperID{4945} 
\def\confName{CVPR}
\def\confYear{2024}

\title{Equivariant Multi-Modality Image Fusion}

\author{Zixiang Zhao$^{1,2}$\quad
        Haowen Bai$^{1}$\quad
        Jiangshe Zhang$^{1}$\quad
        Yulun Zhang$^{3}$\thanks{Corresponding authors.}\quad
        Kai Zhang$^{4}$\\
        Shuang Xu$^{5}$\quad
        Dongdong Chen$^{6\,*}$\quad
        Radu Timofte$^{2,7}$\quad
        Luc Van Gool$^{2,8}$\\[1mm]
        $^{1}$Xi’an Jiaotong University\quad
        $^{2}$ETH Z\"urich\quad
        $^{3}$Shanghai Jiao Tong University\\
        $^{4}$Nanjing University\quad
        $^{5}$Northwestern Polytechnical University\quad
        $^{6}$Heriot-Watt University\\
        $^{7}$University of W\"urzburg\quad
        $^{8}$INSAIT\\
        {\tt\small zixiangzhao@stu.xjtu.edu.cn}
}

\begin{document}
	\maketitle
	
	\begin{abstract}
		Multi-modality image fusion is a technique that combines information from different sensors or modalities, enabling the fused image to retain complementary features from each modality, such as functional highlights and texture details. However, effective training of such fusion models is challenging due to the scarcity of ground truth fusion data. To tackle this issue, we propose the \textbf{E}quivariant \textbf{M}ulti-\textbf{M}odality im\textbf{A}ge fusion (\textbf{EMMA}) paradigm for end-to-end self-supervised learning. Our approach is rooted in the prior knowledge that natural imaging responses are equivariant to certain transformations. Consequently, we introduce a novel training paradigm that encompasses a fusion module, a pseudo-sensing module, and an equivariant fusion module. These components enable the net training to follow the principles of the natural sensing-imaging process while satisfying the equivariant imaging prior. Extensive experiments confirm that EMMA yields high-quality fusion results for infrared-visible and medical images, concurrently facilitating downstream multi-modal segmentation and detection tasks. The code is available at \url{https://github.com/Zhaozixiang1228/MMIF-EMMA}.
	\end{abstract}

	\section{Introduction}\label{sec:1}
	Multi-modality image fusion serves as an image restoration method that synthesizes information from multiple sensors and modalities to generate a comprehensive representation of scenes and objects~\cite{9151265,zhaoijcai2020, Zhao_2023_CVPR,meher2019a}. It finds widespread application in tasks such as image registration~\cite{DBLP:conf/cvpr/Xu0YLL22,DBLP:conf/ijcai/WangLFL22,DBLP:conf/mm/JiangZ0L22}, scene information enhancement or restoration~\cite{DBLP:journals/corr/abs-2104-06977,Liang2022ECCV,fang2024glgnet,yan2022learning,yan2022rignet}, and downstream tasks such as object detection~\cite{DBLP:journals/corr/abs-2004-10934,DBLP:conf/cvpr/LiuFHWLZL22} and semantic segmentation~\cite{DBLP:conf/mm/LiuLL021,DBLP:journals/inffus/TangYM22} in scenes with multiple sensors. Notable tasks include infrared-visible image fusion (IVF) and medical image fusion (MIF). IVF focuses on merging thermal radiation information from input infrared images and intricate texture details from input visible images, resulting in fusion images that mitigate the limitations of visible images affected by illumination variations and infrared images susceptible to low resolution and noise~\cite{10088423,Zhao_2023_ICCV2}.
	The primary goal of MIF is to provide a comprehensive representation of any abnormalities in a patient's medical condition. This is accomplished by integrating multiple imaging techniques, thereby enabling an intelligent decision-making system that supports both diagnostic and therapeutic processes~\cite{DBLP:journals/inffus/JamesD14}.
	
	We assume that the underlying ground truth fused image is information-rich, but in practice we can only measure the same ground truth through different sensing processes which are typically nonlinear and difficult to model, thus obtaining observations in different modalities. Therefore, the multi-modality image fusion problem can be regarded as a challenging \emph{nonlinear and blind} inverse problem, which can be regarded as the following negative log-likelihood minimization problem:
	\begin{subequations}\label{eq:likelihood_minimization}
		\setlength{\abovedisplayskip}{3pt}
		\setlength{\belowdisplayskip}{3pt}
		\begin{align}
			&\min_{\boldsymbol{f}} \{-\log p\left(\boldsymbol{f} \mid \boldsymbol{i}_1,\boldsymbol{i}_2\right)\} \label{eq:likelihood_minimization1}  \\
			\propto &\min_{\boldsymbol{f}} \{-\log p\left(\boldsymbol{i}_1,\boldsymbol{i}_2 \mid \boldsymbol{f}\right)-\log p\left(\boldsymbol{f}\right)\} \label{eq:likelihood_minimization2} \\
			\propto &\min _{\boldsymbol{f}}\{\mathcal{L}(\boldsymbol{f},\boldsymbol{i}_1,\boldsymbol{i}_2)+\mathcal{R}(\boldsymbol{f})\} \label{eq:likelihood_minimization3}
		\end{align}
	\end{subequations}
	where $\boldsymbol{i}_1$, $\boldsymbol{i}_2$, and $\boldsymbol{f}$ represent two input source images and the output fusion image, respectively. \cref{eq:likelihood_minimization2} originates from Bayes' theorem. In \cref{eq:likelihood_minimization3}, the first term is the \textit{data fidelity term}, indicating that 
	$\boldsymbol{i}_1$ and $\boldsymbol{i}_2$ are sensed from $\boldsymbol{f}$;
	the second term is the 
	\textit{prior term}, indicating that $\boldsymbol{f}$ needs to satisfy certain fusion image prior or empirical characteristics.
	
	In the era of deep learning, numerous advanced methods strive to better model this problem. However, several pressing issues remain unaddressed in this task. For the first term of \cref{eq:likelihood_minimization3}, it is evident that individual sensors are limited to capturing modality-specific features; no singular ``super'' sensor exists that can perceive all modal information simultaneously in reality. Consequently, the absence of a definitive ground truth hampers the effective application of deep learning's supervised learning paradigm to image fusion tasks.  
	While generative model-based methods~\cite{ma2019fusiongan,DBLP:conf/cvpr/LiuFHWLZL22} attempt to achieve fusion by making the source image and the fused image belong to a similar distribution, they suffer from a lack of interpretability, controllability, and present training challenges.
	On the other hand, methods based on manually crafted loss functions~\cite{DBLP:journals/inffus/LiWK21,zhaoijcai2020,9151265} often push the fusion image to resemble the source images by minimizing the $\ell_1$ or $\ell_2$ distance. However, such direct computation of $ \|\boldsymbol{f}-\boldsymbol{i}_1\|+ \|\boldsymbol{f}-\boldsymbol{i}_2\|$ to determine $\boldsymbol{f}$ neglects the potential domain differences between the fused images and the source images, failing to consider that $\boldsymbol{f}$ may not reside on the same feature manifold as $\boldsymbol{i}_1$ and $\boldsymbol{i}_2$.
	Meanwhile, for the second term of \cref{eq:likelihood_minimization3}, researchers often presuppose that the fused image exhibits certain structures,
	such as low-rank~\cite{DBLP:journals/tip/LiWK20,DBLP:journals/pami/LiXWLK23}, sparsity~\cite{DBLP:journals/pami/0002D21,DBLP:journals/tip/GaoDXXD22}, multi-scale decomposition~\cite{zhaoijcai2020,Zhao_2023_CVPR}, \etc, and impose priors to restrict the solution space. Nonetheless, due to that ground truth fused images are inaccessible, these priors typically depend on speculative assumptions about the fused images or extrapolations from natural image priors, thereby overly relying on domain knowledge and exhibiting limited adaptability to unseen scenarios.
	
	In response to the challenges mentioned above, we plan to address them from two aspects. First, since aligning distributions and manually crafted loss functions are challenging tasks, we propose to start with the sensing and imaging processes. We aim to learn the sensing, or say, the inverse mapping from the fusion image back to images of various modalities. This approach is intuitively simpler than mastering the process of fusion itself. By doing so, we can measure the loss between the input source images and the (pseudo) sensing results, which are obtained by applying the fusion images to different sensing functions. This strategy overcomes the problem of not having ground truth images for fusion. Furthermore, as image fusion is an inherently ill-posed problem, merely optimizing the aforementioned sensing loss may not yield the optimum fused image. Consequently, we introduce a conceptually simple yet effective prior, which is based on the inherent priors of the imaging systems and does not rely on domain-specific knowledge of fusion images. 
	This non-domain-specific prior is predicated on the understanding that natural imaging responses are equivariant to transformations such as shifts, rotations, and reflections. 
	In other words, the transformed fused image, after sensing and re-fused, should yield the same outcome as before sensing. 
	Leveraging the equivariance prior of the natural imaging system  offers stronger constraints and guidance for the learning process within the fusion network. In summary, regarding the common learning paradigms for image fusion, we have made the following improvements:
	\begin{equation}\label{eq:improvements}
		\setlength{\abovedisplayskip}{3pt}
		\setlength{\belowdisplayskip}{3pt}
		\begin{aligned}
			& \|\boldsymbol{f}-\boldsymbol{i}_1 \|\!+\!\|\boldsymbol{f}-\boldsymbol{i}_2 \|\!+\!\text{fusion\ image\ prior}(\boldsymbol{f})\\
			\Longrightarrow  
			& \|\boldsymbol{\hat{i}}_1-\boldsymbol{i}_1 \|\!+\!\|\boldsymbol{\hat{i}}_2-\boldsymbol{i}_2 \|\!+\!\text{equivariance\ prior}(\mathcal{F}\!\circ\!\mathcal{A})
		\end{aligned}
	\end{equation}
	where $\mathcal{F}$ represents the fusion model and $\mathcal{A}$ is the sensing model. $\boldsymbol{\hat{i}}_1\!=\!\mathcal{A}_1(\boldsymbol{f})$ and $\boldsymbol{\hat{i}}_2\!=\!\mathcal{A}_2(\boldsymbol{f})$ denote the respective sensing results for $\boldsymbol{i}_1$ and $\boldsymbol{i}_2$, as determined by their corresponding sensing models $\mathcal{A}_1$ and $\mathcal{A}_2$, respectively. $\mathcal{A}_1$ and $\mathcal{A}_2$ together comprise the sensing model $\mathcal{A}$.
	
	Following this methodology, we devise a self-supervised learning paradigm named \textit{\textbf{E}quivariant \textbf{M}ulti-\textbf{M}odality im\textbf{A}ge fusion} (\textbf{EMMA}).
	This framework consists of a fusion module, a pseudo-sensing module, and an equivariant fusion module. The fusion module, named U-Fuser, is a U-Net-like~\cite{DBLP:conf/miccai/RonnebergerFB15} structure that incorporates  Restormer~\cite{DBLP:conf/cvpr/ZamirA0HK022}-CNN blocks, and is employed to model both global and local features, thereby effectively aggregating information. The pseudo-sensing module, based on U-Net~\cite{DBLP:conf/miccai/RonnebergerFB15}, is a learnable construct that maps the fused image back to the source images, simulating the natural process of perception imaging. Lastly, the equivariant fusion module is designed to ensure that fused images adhere to the established prior of equivariant imaging. Our contributions are as follows:
	\begin{itemize}
		\item We propose a novel self-supervised learning paradigm named EMMA, designed to address the absence of ground truth in image fusion. EMMA leverages the natural sensing-imaging process with the non-domain-specific prior that imaging responses are equivariant to transformations such as shift, rotation, and reflection.
		\item We refine the inappropriate handling of domain differences between fused images and source inputs in the conventional fusion loss by simulating the perceptual imaging process via pseudo-sensing module and the sensing loss component effectively.
		\item The U-Fuser fusion module proposed in EMMA proficiently models long- and short-range dependencies across multiple scales to integrate the source information.
		\item Our approach demonstrates excellent performance in infrared-visible image fusion and medical image fusion, which is also proved to facilitate downstream multi-modal object detection and semantic segmentation tasks.
	\end{itemize}

	\section{Related Work}\label{sec:2}
	\bfsection{Multi-modality image fusion}
	In the deep learning era, multi-modality image fusion methods can be classified into four primary groups: generative models~\cite{ma2019fusiongan,ma2020infrared,DBLP:journals/tip/MaXJMZ20}, autoencoder-based models~\cite{li2018densefuse,DBLP:conf/mm/LiuLL021,DBLP:journals/inffus/LiWK21,DBLP:journals/ijcv/ZhangM21,liu2023coconet}, algorithm unrolling models~\cite{DBLP:journals/pami/0002D21,DBLP:journals/tip/GaoDXXD22,DBLP:journals/tcsv/ZhaoXZLZL22,DBLP:journals/corr/abs-2005-08448}, and unified models~\cite{xu2020aaai,DBLP:conf/aaai/ZhangXXGM20,9151265,DBLP:journals/inffus/ZhangLSYZZ20,DBLP:journals/tip/JungKJHS20}. Generative models represent the distribution of fused images and source images in the latent space through generative adversarial networks~\cite{ma2019fusiongan,ma2020infrared,DBLP:journals/tip/MaXJMZ20} or denoising diffusion model~\cite{Zhao_2023_ICCV}. 
 	Autoencoder-based models use the encoder/decoder with CNN or Transformer block as the basic unit to model the mapping/inverse mapping between the image domain and the feature domain~\cite{Zhao_2023_CVPR,Liang2022ECCV,vs2022image}. 
 Algorithm unrolling models shift the algorithm focus from data-driven learning to model-driven learning, which {replace} complex operators with CNN/Transformer blocks while retaining the original computational graph structure, achieving lightweight and interpretable learning~\cite{DBLP:journals/tcsv/ZhaoXZLZL22,DBLP:journals/pami/LiXWLK23}. 
 Unified models identify meta-knowledge between different tasks through cross-task learning, enabling rapid adaptation to new tasks and improved performance with fewer examples~\cite{9151265,DBLP:journals/ijcv/ZhangM21}.
	Moreover, the multi-modality image fusion task is often integrated into coupled systems with upstream (pre-processing) image registration~\cite{DBLP:conf/cvpr/Xu0YLL22,DBLP:conf/ijcai/WangLFL22,DBLP:conf/mm/JiangZ0L22} and downstream object detection and semantic segmentation tasks~\cite{DBLP:conf/cvpr/LiuFHWLZL22,DBLP:journals/inffus/TangYM22,DBLP:conf/mm/SunCZH22,Liu_2023_ICCV}. Image registration can effectively eliminate image artifacts and unaligned areas, enhance edge clarity and expand the perception field~\cite{DBLP:conf/cvpr/Xu0YLL22,huangreconet,DBLP:journals/pami/XuYM23}. Furthermore, gradient of the recognition loss in downstream tasks can effectively guide the production of the fused image \cite{DBLP:conf/cvpr/LiuFHWLZL22,DBLP:journals/inffus/TangYM22,DBLP:conf/cvpr/ZhaoXZHL23,Liu_2023_ICCV}.
	
	\bfsection{Equivariant Imaging}
	Equivariant imaging (EI) \cite{DBLP:journals/spm/ChenDESST23,DBLP:conf/cvpr/0004TD22,DBLP:conf/iccv/0004TD21} is an emerging fully unsupervised imaging framework that exploits the group invariance property in natural signals to learn a reconstruction function from partial measurement data alone. 
	The main idea behind EI is to use the fact that natural signals often have certain symmetries. For example, images are often translation invariant, meaning that they look the same if they are shifted around. With this invariance
	prior, the whole imaging system (from sensing to reconstruction) is transformation equivariant.
	Under certain sensing conditions \cite{JMLR:v24:22-0315}, the reconstruction function will be able to correctly reconstruct images that have been transformed around, even if it has never seen those images before. As a promising new approach to imaging and a new way to acquire and process images, EI has been shown to be effective for a variety of linear inverse problems~\cite{DBLP:journals/spm/ChenDESST23}. This paper devotes to exploring the potential of EI on a more challenging task, \ie, non-linear and blind inverse problems in multi-modality image fusion.
	
	\bfsection{Comparison with existing approaches}
	\textbf{a)} Compared to the regular fusion loss, \ie $\|\boldsymbol{f}-\boldsymbol{i}_1\|+ \|\boldsymbol{f}-\boldsymbol{i}_2\|$ in the image or feature domains~\cite{DBLP:journals/inffus/LiWK21,zhaoijcai2020,DBLP:journals/inffus/TangYM22}, the pseudo-sensing loss item in \cref{eq:improvements} from EMMA {mitigates} the irrationality in traditional loss caused by the manifold difference between $\boldsymbol{f}$ and $\{\boldsymbol{i}_1,\boldsymbol{i}_2\}$, ensuring that the distances calculated between 
	$\{\boldsymbol{\hat{i}}_1,\boldsymbol{i}_1\}$ and  $\{\boldsymbol{\hat{i}}_2,\boldsymbol{i}_2\}$ are within the same domain.
	\textbf{b)} Similar fusion-to-source mapping concepts \cite{DBLP:journals/ijcv/ZhangM21,ye2023lfienet} aim to make $\boldsymbol{f}$ decomposable into $\{\boldsymbol{i}_1,\boldsymbol{i}_2\}$ to ensure it containing the source image information. However, their decomposition module, as an integral part of the fusion algorithm, undergoes updates during training, and the fusion output is considered as a feature for source reconstruction. Thus, proficiency in decomposition learning does not invariably correlate with enhanced information in fusion. In contrast, within the EMMA paradigm, the learning of the pseudo-sensing module is decoupled from that of the fusion network, and it remains frozen during EMMA training, thus ensuring that the mapping from the fused image back to the source image is explicit and determinate. This enhances the rationality and interpretability of the sensing module.
	\textbf{c)} Furthermore, other prior-based optimizations~\cite{zhaoijcai2020,DBLP:journals/pami/LiXWLK23} often necessitate domain knowledge of fusion images. However, in EMMA, we only need to use the 
	imaging system 
	prior rather than the fusion image prior to accomplish self-supervised learning.

	\section{Method}
	In this section, we first provide the model formalization, including the sensing module and the fusion module, and give the model hypotheses for establishing the equivariant image fusion paradigm.
	Then, we take the IVF task as an example and present the implementation details of EMMA. Other image fusion tasks can be analogously derived.
	\subsection{Problem Overview}
	Let
	$\boldsymbol{i}$, $\boldsymbol{v}$, and $\boldsymbol{f}$ refer to infrared, visible, and fused images, respectively, with $\boldsymbol{i}\!\in\!\mathbb{R}^{HW}$, $\boldsymbol{v}\!\in\!\mathbb{R}^{3HW}$, and $\boldsymbol{f}\!\in\!\mathbb{R}^{3HW}$.
	We assume the existence of an information-rich $\boldsymbol{f}$ that contains multi-sensory and multi-modal information and needs to be predicted. However, there is no perception device in real life that can fully sense $\boldsymbol{f}$ up to now. Thus, as an unsupervised task, there is no ground truth for $\boldsymbol{f}$. Therefore, we model the fusion process and the sensing process as follows:
	\begin{equation}\label{}
		\setlength{\abovedisplayskip}{3pt}
		\setlength{\belowdisplayskip}{3pt}   \boldsymbol{f}\!=\!\mathcal{F}\left(\boldsymbol{i},\boldsymbol{v}\right)\!+\!\boldsymbol{n}_f\Leftrightarrow
		\boldsymbol{i}\!=\!\mathcal{A}_i\left(\boldsymbol{f}\right)\!+\!\boldsymbol{n}_i, \boldsymbol{v}\!=\!\mathcal{A}_v\left(\boldsymbol{f}\right)\!+\!\boldsymbol{n}_v,
	\end{equation}
	where $\mathcal{F}\left(\cdot,\cdot \right) $ represents the fusion model, $\mathcal{A}_i\left(\cdot\right)$ and $\mathcal{A}_v\left(\cdot\right)$ represent the sensing model of $\boldsymbol{i}$ and $\boldsymbol{v}$, \emph{i.e.}, the infrared and RGB cameras, respectively.
	In the traditional image inverse problem $\boldsymbol{y}=\mathcal{A}(\boldsymbol{x})+\boldsymbol{n}$, where $\boldsymbol{x}$ and $\boldsymbol{y}$ are the ground truth image and the measurement, the degradation operator $\mathcal{A}(\cdot)$ is known (such as the noise distribution in denoising tasks and the blur kernel in super-resolution tasks). However, in image fusion, we cannot explicitly obtain $\mathcal{A}_i$ and $\mathcal{A}_v$. 
 Nevertheless, we can set them learnable, in order to simulate the perceptual process and assist the network in self-supervised learning.
	\subsection{Model hypothesis}\label{sec:hypothesis}
	To provide comprehensive sensing and fusion models and further support the subsequent introduction of EMMA framework, we first need to establish some necessary hypotheses.
 
	\bfsection{a) Measurement consistency}
	We assume that the fusion function $\mathcal{F}\left(\cdot,\cdot \right)$ maintains consistency within the measurement domain, that is,
	\begin{equation}\label{eq:fusion_model}
		\setlength{\abovedisplayskip}{3pt}
		\setlength{\belowdisplayskip}{3pt}	\mathcal{A}_i\left(\mathcal{F}\left(\boldsymbol{i},\boldsymbol{v}\right)\right) = \boldsymbol{i}, \
		\mathcal{A}_v\left(\mathcal{F}\left(\boldsymbol{i},\boldsymbol{v}\right)\right) = \boldsymbol{v}.
	\end{equation}
	However, due to the underdetermined nature of the sensing process, the estimation of $\mathcal{F}\left(\boldsymbol{i},\boldsymbol{v}\right)$ cannot be achieved by estimating the inverse of $\mathcal{A}_i$ or $\mathcal{A}_v$, and we have to learn more information beyond the range space of their inverse.

	\bfsection{b) Invariant set consistency} We first give two definitions in the equivariant imaging \cite{DBLP:journals/spm/ChenDESST23}:
        \vspace{-0.5em}
	\begin{definition}[Invariant set]\label{definition1}
		For a set of transformations $\mathcal{G}=\left\{g_1, \ldots, g_{|\mathcal{G}|}\right\}$ composed of unitary matrices $T_g \in \mathbb{R}^{n \times n}$,
		$\mathcal{X}$ is the invariant set with respect to transformations $\mathcal{G}$, if $T_g x \in \mathcal{X}$ holds for $ \forall x \in \mathcal{X}$ and $\forall g \in \mathcal{G}$, i.e., {$T_g \mathcal{X}$ and $\mathcal{X}$ are identical}. 
	\end{definition}
 
	\begin{definition}[Equivariant function]\label{definition2}
		If function $\mathcal{I}$ satisfies $\mathcal{I}\left(T_g x\right) = T_g \mathcal{I}(x)$ for $\forall x \in \mathcal{X}$ and $\forall g \in \mathcal{G}$, we call 
		$\mathcal{I}$ is an equivariant function with respect to the transformation $\mathcal{G}$.
	\end{definition}
	Regarding the corollary of \cref{definition1}, if $\mathcal{X}$ represents a set of natural images, it is evident that the result remains natural images after transformations that include translations, rotations, and reflections. Hence, $\mathcal{X}$ is an invariant set for transformation group $\mathcal{G}$. Furthermore, the set composed of fused images $\boldsymbol{f}$, being a subset of $\mathcal{X}$, is also an invariant set to $\mathcal{G}$.
	Moreover, in \cref{definition1,definition2}, ``invariance'' pertains to the properties of the dataset, while ``equivariance'' characterizes the properties of the imaging system, meaning that the imaging system (denoted as $\mathcal{F}\!\circ\!\mathcal{A}$ in our paper) is the equivariant function with respect to $\mathcal{G}$.
	Consequently, we propose the following theorem:
	\begin{theorem}[Equivariant image fusion theorem]\label{fusiontheorem}
		If we regard $\mathcal{I}$ in \cref{definition2} to be the composite function $\mathcal{F}\!\circ\!\mathcal{A}$, where $\mathcal{F}$ is the fusion model and $\mathcal{A}$ (including $\mathcal{A}_i$ and $\mathcal{A}_v$) is the sensing model, the {equivariant image fusion theorem} is:
		\begin{equation}\label{eq:equivariant_fusion_hypothesis}        	
            \setlength{\abovedisplayskip}{3pt}
            \setlength{\belowdisplayskip}{3pt} 
            \mathcal{F}\left(\mathcal{A}_i\left(T_g\boldsymbol{f}\right),\mathcal{A}_v\left(T_g\boldsymbol{f}\right) \right)=T_g\mathcal{F}\left(\mathcal{A}_i\left(\boldsymbol{f}\right),\mathcal{A}_v\left(\boldsymbol{f}\right)\right).
		\end{equation}
	\end{theorem}
	
	\renewcommand\qedsymbol{$\blacksquare$}
	\begin{proof}
		Consider a set of natural images $\mathcal{X}$ satisfying the invariance property, by \cref{definition2} the imaging system $\mathcal{F}\circ \mathcal{A}$ should be equivariant to the group actions $\{T_g\}$. Hence, for $\forall \boldsymbol{f} \in \mathcal{X}$, we have $\mathcal{F}\circ \mathcal{A}(T_g \boldsymbol{f})=T_g \mathcal{F}\circ \mathcal{A}(\boldsymbol{f})$. Furthermore, by separating $\mathcal{A}$ into $\mathcal{A}_i$ and $\mathcal{A}_v$, we can get \cref{eq:equivariant_fusion_hypothesis}.
	\end{proof}
	
	\begin{remark}
		For \cref{eq:equivariant_fusion_hypothesis}, it does not necessitate $\mathcal{F}$ or $\mathcal{A}$ to be equivariant to $T_g$, instead, $\mathcal{F} \circ \mathcal{A}$ is required to be equivariant. Thus, $\mathcal{F}$ and $\mathcal{A}$ here can be set to any form of mapping without restriction.
	\end{remark}
	In the following 
	, we will demonstrate how to establish our 
	equivariant image
	fusion
	paradigm based on \cref{fusiontheorem}.

	\begin{figure*}[t]
		\centering
		\includegraphics[width=0.82\linewidth]{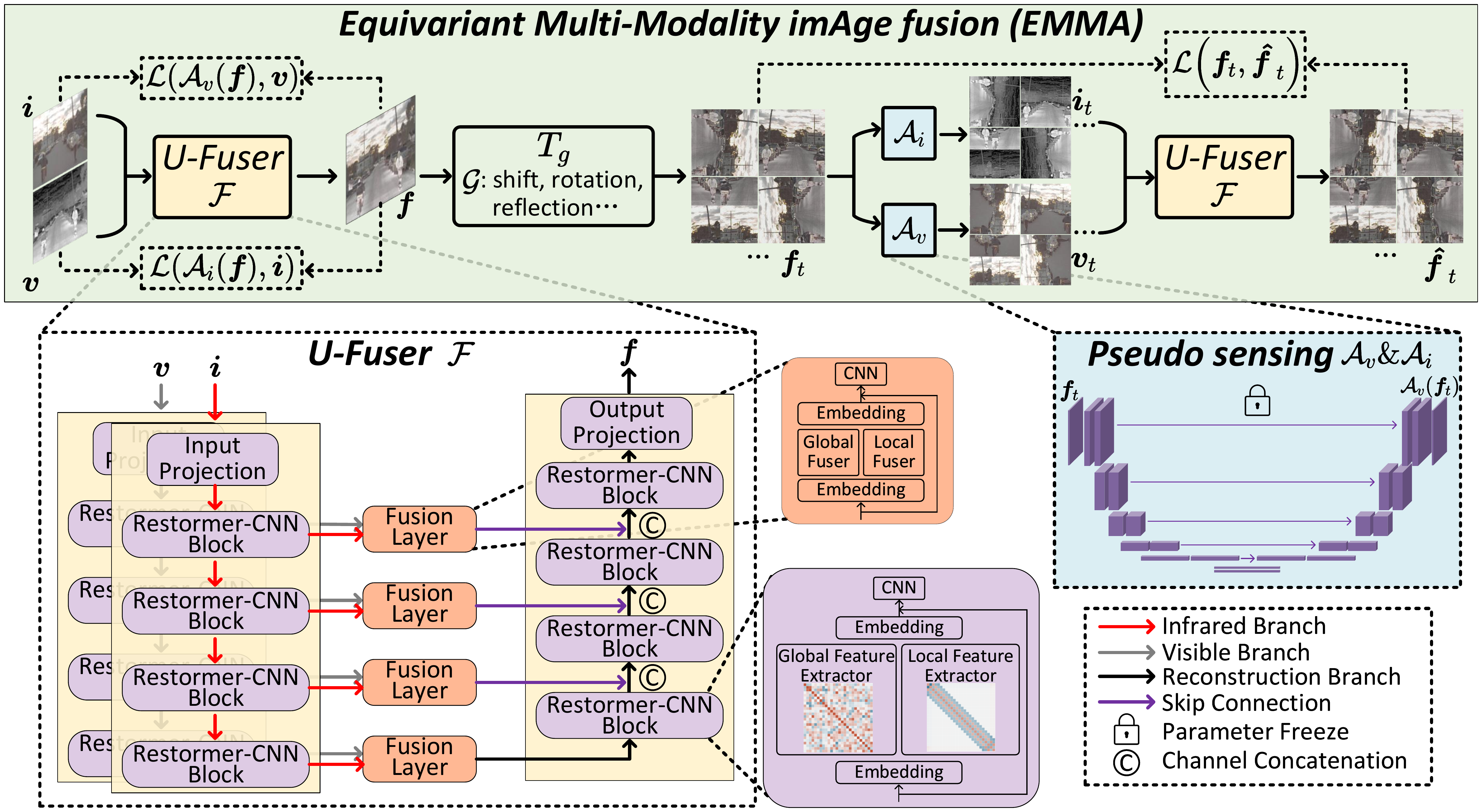}
		\caption{Workflow for EMMA. The image pair $\left\{\boldsymbol{i},\boldsymbol{v}\right\}$ are initially input into U-Fuser $\mathcal{F}$, resulting in the fused image $\boldsymbol{f}$.
			Next, a series of transformations $T_g$ containing shift, rotation, reflection, \emph{etc.}, are applied to $\boldsymbol{f}$ to produce $\boldsymbol{f}_t$.
			$\boldsymbol{f}_t$ is then fed into the parameter-frozen $\left\{\mathcal{A}_i,\mathcal{A}_v\right\}$ to generate the pseudo-sensing images $\left\{\boldsymbol{i}_t,\boldsymbol{v}_t\right\}$, which are finally input into $\mathcal{F}$ to obtain the re-fused image $\boldsymbol{\hat{f}}_t$.}
		\label{fig:Workflow}
		\vspace{-1.5em}
	\end{figure*}
	
	\subsection{Equivariant image fusion paradigm}
	The main focus of this paper is to present EMMA, a self-supervised image fusion framework based on the equivariant imaging prior, with the specific workflow shown in \cref{fig:Workflow}.
	
	\bfsection{Overall paradigm}
	Firstly, we establish a U-Net-like fusion module $\mathcal{F(\cdot)}$ named \textit{U-Fuser}, which combines a Restormer~\cite{DBLP:conf/cvpr/ZamirA0HK022} with CNN blocks as the basic unit to generate the fused image $\boldsymbol{f}$ from inputs $\boldsymbol{i}$ and $\boldsymbol{v}$.
	Subsequently, based on the equivariant image fusion theorem in \cref{fusiontheorem}, an equivariant prior-based self-supervised framework, comprising \textit{{U-Fuser} module} and learnable \textit{(pseudo) sensing modules} $\mathcal{A}_i$ and $\mathcal{A}_v$, is employed to better preserve the source image information in the absence of the fusion ground truth.
	Specifically, we transform $\boldsymbol{f}$, estimated by U-Fuser, through a series of transformations $T_g$ to obtain $\boldsymbol{f}_t$, then pass $\boldsymbol{f}_t$ through pseudo sensing modules $\left\{\mathcal{A}_i,\mathcal{A}_v\right\}$ to obtain pseudo-images $\left\{\boldsymbol{i}_t,\boldsymbol{v}_t\right\}$. Finally, we fuse $\left\{\boldsymbol{i}_t,\boldsymbol{v}_t\right\}$ with U-Fuser again to obtain $\boldsymbol{\hat{f}}_t$.
	
	Unlike other methods that require a well-designed loss function to minimize the distance between $\boldsymbol{f}$ and $\left\{\boldsymbol{i},\boldsymbol{v}\right\}$, EMMA's loss focuses on making the pseudo-images $\left\{\mathcal{A}_i\!\left(\boldsymbol{f}\right),\mathcal{A}_v\!\left(\boldsymbol{f}\right)\right\}$ generated by the sensing module from $\boldsymbol{f}$ as close to the original $\left\{\boldsymbol{i},\boldsymbol{v}\right\}$ as possible, while making $\boldsymbol{f}_t$ close to $\boldsymbol{\hat{f}}_t$ simultaneously. Thus, from a natural imaging perspective, the optimal fusion image $\boldsymbol{f}$ is found.
	
	In the following text, we will first introduce the fusion module U-Fuser $\mathcal{F(\cdot)}$ and the pseudo sensing modules $\left\{\mathcal{A}_i,\mathcal{A}_v\right\}$, then illustrate the entire self-supervised learning framework, and finally provide the training loss function.
	
	\bfsection{U-Fuser module}
	We adopt a U-Net-like structure for fusing $\boldsymbol{i}$ and $\boldsymbol{v}$ and generating the fused image $\boldsymbol{f}$.
	At each scale, since the input cross-modal features contain both global features such as environment and background information, as well as local features like the highlighting and detailed texture object features, we design a Transformer-CNN structure to better model the cross-modal features by leveraging their respective inductive biases.	
	For the selection of Transformer block, we adopt Restormer block~\cite{DBLP:conf/cvpr/ZamirA0HK022}, which implements self-attention in channel dimension to model global features without too much computation load. In the CNN block, we use Res-block~\cite{HeZRS16}. 
	The input features of the Restormer-CNN block are embedded and then parallelly processed by the Restormer block and the Res-block, followed by embedding interaction and a CNN layer, and finally input to the next scale.
	Features of $\boldsymbol{i}$ and $\boldsymbol{v}$ at the same scale are fused in the fusion layer, and are passed to the reconstruction branch at the previous scale via skip connections.
	Design of blocks for feature fusion and reconstruction is the same as Restormer-CNN block used in the feature extraction branch.
	
	\bfsection{Pseudo sensing module}\label{sec:pseudo_sensing}
	In contrast to other works in this field where their algorithm mainly focuses on the design of the fusion function $\mathcal{F}$, in this paper, we propose a self-supervised learning framework based on equivariant imaging prior to address the issue of lacking ground truth for fused images. According to the equivariant image fusion theorem stated in \cref{fusiontheorem}, we need to obtain pseudo imaging results from $\mathcal{A}_i\!\left(\boldsymbol{f}\right)$ and $\mathcal{A}_v\!\left(\boldsymbol{f}\right)$. To achieve this goal, we need to simulate the process of sensing infrared and visible images from the (imagined) fused image, as described in \cref{eq:fusion_model}. Since it is not feasible to explicitly give the structures of $\mathcal{A}_i$ and $\mathcal{A}_v$, we adopt a data-driven learning approach to obtain them. Recently, many deep learning-based methods have shown promising results in image fusion. Therefore, we selected fifteen state-of-the-art (SOTA) methods that have recently appeared in top venues. They are DIDFuse~\cite{zhaoijcai2020}, U2Fusion~\cite{9151265}, SDNet~\cite{DBLP:journals/ijcv/ZhangM21}, RFN-Nest~\cite{DBLP:journals/inffus/LiWK21}, AUIF~\cite{DBLP:journals/tcsv/ZhaoXZLZL22}, RFNet~\cite{DBLP:conf/cvpr/Xu0YLL22}, TarDAL~\cite{DBLP:conf/cvpr/LiuFHWLZL22}, DeFusion~\cite{Liang2022ECCV},  ReCoNet~\cite{huangreconet},
	MetaFusion~\cite{DBLP:conf/cvpr/ZhaoXZHL23},
	CDDFuse~\cite{Zhao_2023_CVPR},
	LRRNet~\cite{DBLP:journals/pami/LiXWLK23},
	MURF~\cite{DBLP:journals/pami/XuYM23},
	DDFM~\cite{Zhao_2023_ICCV} and
	SegMIF~\cite{Liu_2023_ICCV}. We use their fusion results as the (pseudo) ground truth for the fused images and then learn the mappings from the fused images to $\boldsymbol{i}$ and $\boldsymbol{v}$, which can be regarded as $\mathcal{A}_i$ and $\mathcal{A}_v$, respectively. Considering that both the input and output of the mapping have the same image size, we choose U-Net~\cite{DBLP:conf/miccai/RonnebergerFB15} as the backbone of  $\mathcal{A}_i$ and $\mathcal{A}_v$ and conduct the end-to-end training paradigm. The specific network details are in the supplementary material.
 
	\bfsection{Equivariant image fusion}
	After obtaining the U-Fuser $\mathcal{F}$ and pseudo-sensing functions $\left\{\mathcal{A}_i,\mathcal{A}_v\right\}$, we introduce our self-supervised learning framework based on image equivariant prior. As shown in \cref{fig:Workflow}, we first input the image pairs $\left\{\boldsymbol{i},\boldsymbol{v}\right\}$ into $\mathcal{F}$, and obtain fused image $\boldsymbol{f}$ (which is the entire operation of conventional fusion algorithms).
	Then, we apply a series of transformations $T_g$ to $\boldsymbol{f}$, including shift, rotation, reflection, \etc, to obtain $\boldsymbol{f}_t$.
	Subsequently, $\boldsymbol{f}_t$ is input into the well-trained $\left\{\mathcal{A}_i,\mathcal{A}_v\right\}$ to obtain the pseudo-sensing images $\left\{\boldsymbol{i}_t,\boldsymbol{v}_t\right\}$, which contain the information from $\boldsymbol{f}_t$ and satisfy the imaging characteristics of infrared and visible images, respectively.
	Finally, paired $\left\{\boldsymbol{i}_t,\boldsymbol{v}_t\right\}$ are fed into $\mathcal{F}$ to obtain the re-fused image $\boldsymbol{\hat{f}}_t$. Throughout the framework, we aim to aggregate information from $\left\{\boldsymbol{i},\boldsymbol{v}\right\}$ into $\boldsymbol{f}$, and according to the equivariant image fusion theorem (\cref{fusiontheorem}), $\boldsymbol{f}_t$ and $\boldsymbol{\hat{f}}_t$ should be sufficiently close. These will be guaranteed through the designed loss function.
 
 	\bfsection{Training detail and loss function}
	During the entire training process of EMMA, we first trained  $\mathcal{A}_i$ and $\mathcal{A}_v$ using $\ell_2$ loss as the loss function, \emph{i.e.}, $\mathcal{L}_{I}^{Rec}=\ell_2\left(\boldsymbol{i},\mathcal{A}_i(\tilde{\boldsymbol{f}})\right)$ and $\mathcal{L}_{V}^{Rec}=\ell_2\left(\boldsymbol{v},\mathcal{A}_v(\tilde{\boldsymbol{f}})\right)$, where $\tilde{\boldsymbol{f}}$ are the fusion results from the SOTA methods in \cref{sec:pseudo_sensing}. Then, we freeze the parameters of $\mathcal{A}_i$ and $\mathcal{A}_v$, which means that parameters of the pseudo-sensing module will no longer be updated. Afterwards, we train U-Fuser module with the total loss function:
	\begin{equation}\label{eq:total_loss}
		\small
		\setlength{\abovedisplayskip}{3pt}
		\setlength{\belowdisplayskip}{3pt} 
        \mathcal{L}_{total}\!=\!\mathcal{L}\left(\mathcal{A}_i\!\left(\boldsymbol{f}\right)\!,\!\boldsymbol{i}\right)
		+\alpha_1 \mathcal{L}\left(\mathcal{A}_v\!\left(\boldsymbol{f}\right)\!,\!\boldsymbol{v}\right)
		+\alpha_2 \mathcal{L}\left(\boldsymbol{f}_t, \boldsymbol{\hat{f}}_t\right),
	\end{equation}
	where {$\mathcal{L}(\boldsymbol{x},\hat{\boldsymbol{x}}) =\ell_1(\boldsymbol{x},\hat{\boldsymbol{x}})+\ell_1(\nabla\boldsymbol{x},\nabla\hat{\boldsymbol{x}})$}. $\alpha_1$ and $\alpha_2$ are the tuning parameters, and {$\nabla$ indicates the Sobel  operator}. In particular, the first and second terms of \cref{eq:total_loss} ensure our paradigm satisfies the \textit{measurement consistency} of model hypothesis in \cref{sec:hypothesis}, while the third term ensures it satisfies the \textit{invariant set consistency} of model hypothesis.
	
	\begin{figure*}[t]
	\centering
	\includegraphics[width=0.9\linewidth]{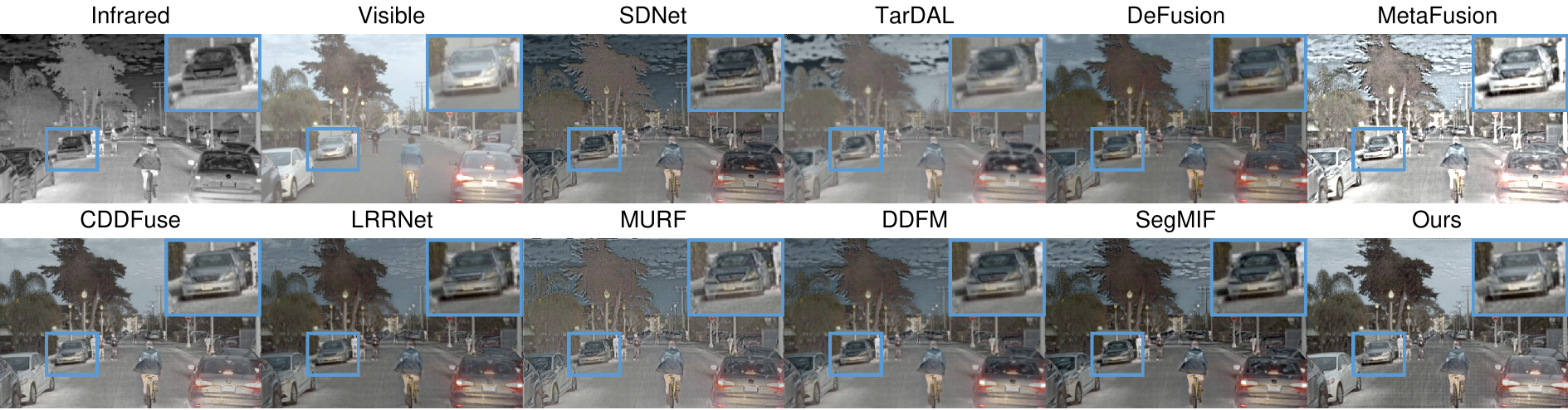}
	\caption{Visual comparison of ``06832'' from RoadScene~\cite{xu2020aaai} IVF dataset.}
	\label{fig:IVF1}
	\vspace{-1em}
\end{figure*}
\begin{figure*}[t]
	\centering
	\includegraphics[width=0.9\linewidth]{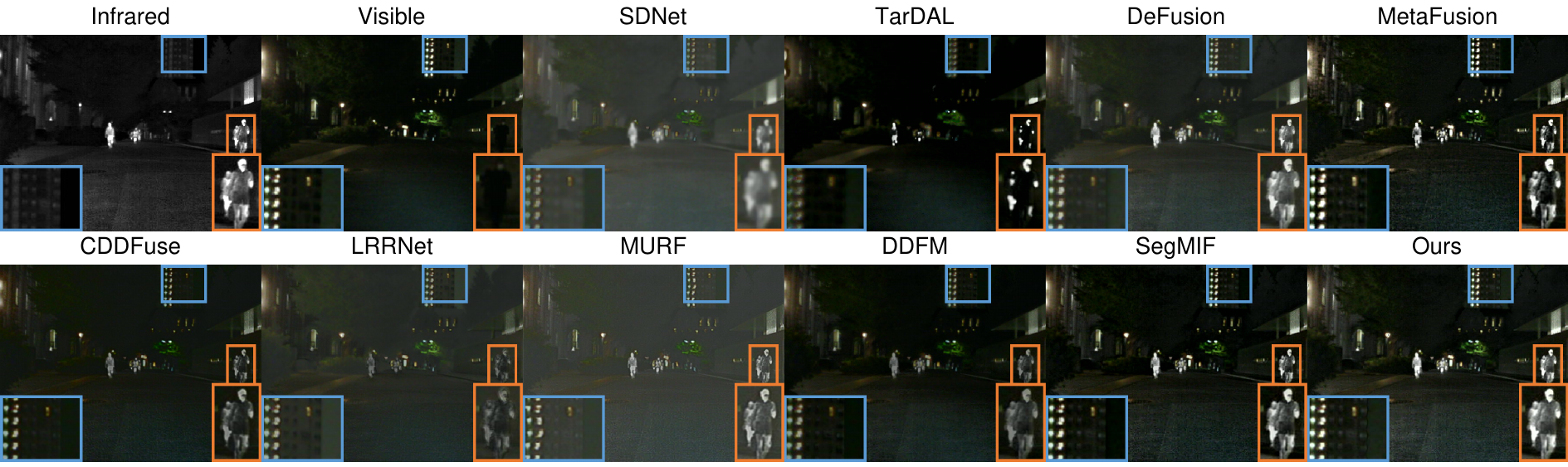}
	\caption{Visual comparison of ``00782N'' from MSRS~\cite{DBLP:journals/inffus/TangYZJM22} IVF dataset.}
	\label{fig:IVF2}
	\vspace{-1.5em}
\end{figure*}

	\subsection{Explanations}

Here we will explain why the unsupervised fusion of EMMA works. By the fact that image set $\{\boldsymbol{f}\}$ is invariant to a group of invertible transformations $\{T_g\}$, give any image $\boldsymbol{f}$ from the invariant set $\{\boldsymbol{f}\}$, then $T_g\boldsymbol{f}$ also belongs to the set for all $g=1,\cdots, |G|$. Under the equivariant theorem in \cref{fusiontheorem}, we have $\{\boldsymbol{i},\boldsymbol{v}\}=\mathcal{A}\boldsymbol{f}=\mathcal{A}T_gT_g^{-1}\boldsymbol{f}=\mathcal{A}_g\boldsymbol{f}'$ for $g=1,\cdots,|G|$, where $\mathcal{A}_g=\mathcal{A}T_g$ and $\boldsymbol{f}'=T_g^{-1}\boldsymbol{f}$ belongs to $\{\boldsymbol{f}\}$.
That is to say, applying transformations is equal to generating multiple virtual sensing operators $\{\mathcal{A}_{g}\}_{g=1,\cdots,|G|}$. Since those virtual operators $\mathcal{A}_g$ are with potentially different nullspaces, this allows us to learn beyond the range space of inverse $\mathcal{A}$ (see \cite{JMLR:v24:22-0315}). 

The lack of ground truth leads to potential inaccuracies in modeling $\mathcal{A}_i$ and $\mathcal{A}_v$, making the reconstruction of $\boldsymbol{f}$ potentially unsatisfactory in the first few training epochs.
Fortunately, the combination of transformation for $\boldsymbol{f}_t$ and learning via equivariant imaging prior allows the completion of the originally missing knowledge to calibrate and refine the fusion results, \ie, achieving the recovering of the missed null space component. Notably, in the final algorithm deployment phase, only the fine-tuned U-Fuser $\mathcal{F}$ is needed, and all other modules will be disregarded, such as $\mathcal{A}_i$ and $\mathcal{A}_v$.  Finally, the proposed equivariant fusion module differs from data augmentation (DA), which mainly extends data based on the ground truth. However, ground truth is firmly inaccessible in the image fusion task 
and DA cannot provide extra information gains when learning to image without ground truth \cite{DBLP:journals/spm/ChenDESST23,DBLP:conf/iccv/0004TD21}. 
Fortunately, as we have shown, with the equivariance prior the proposed EMMA can provide extra information and figure out principle-plausible fusion results.

	\begin{table*}[t]
		\centering
		\resizebox{0.85\linewidth}{!}{
			\begin{tabular}{cccccccccccccc}
				\toprule
				\multicolumn{7}{c}{\textbf{Infrared-Visible Image Fusion on MSRS Dataset}~\cite{DBLP:journals/inffus/TangYZJM22}}                              &                              \multicolumn{7}{c}{\textbf{Infrared-Visible Image Fusion on RoadScene Dataset}~\cite{xu2020aaai}}                               \\
			&  EN $\uparrow$   &   SD $\uparrow$   &  SF $\uparrow$   &  AG $\uparrow$  & SCD $\uparrow$  & VIF  $\uparrow$  &                                        &  EN $\uparrow$   &   SD $\uparrow$   &  SF $\uparrow$   &  AG $\uparrow$  & SCD $\uparrow$  & VIF  $\uparrow$  \\ \midrule
			SDN~\cite{DBLP:journals/ijcv/ZhangM21}  & 5.25  & 17.35  & 8.67  & 2.67  & 0.99  & 0.50  & SDN~\cite{DBLP:journals/ijcv/ZhangM21}  & 7.30  &  44.06 & 14.58  & 5.80  &  1.37 & 0.61  \\
			TarD~\cite{DBLP:conf/cvpr/LiuFHWLZL22}  & 5.28  & 25.22  & 5.98  & 1.83  & 0.71  & 0.42  & TarD~\cite{DBLP:conf/cvpr/LiuFHWLZL22} & 7.26  & 47.44  & 11.11  & 4.14  & 1.40  &  0.56 \\
			 DeF~\cite{Liang2022ECCV}  & 6.46  & 37.63  & 8.60  & 2.80  & 1.35  & 0.77  &  DeF~\cite{Liang2022ECCV}  & 7.36  & 47.03 & 10.99  & 4.38  & 1.62  & 0.63  \\
			Meta~\cite{DBLP:conf/cvpr/ZhaoXZHL23}  & 5.65  & 24.97  &  9.99 & 3.40  & 1.14  & 0.31  & Meta~\cite{DBLP:conf/cvpr/ZhaoXZHL23} & 6.88  & 31.97  & 14.38 & 5.57 & 0.92 & 0.55 \\
			CDDF~\cite{Zhao_2023_CVPR} & \underline{6.70}  & \underline{43.38}  & \underline{11.56}  & \underline{3.73}  & \underline{1.62}& \textbf{1.05}  & CDDF~\cite{Zhao_2023_CVPR} & \underline{7.52}  & \underline{54.42}  & \underline{14.97} &  \underline{5.81} & \underline{1.65}  & \underline{0.66}  \\
			LRR~\cite{DBLP:journals/pami/LiXWLK23} & 6.19  & 31.78  & 8.46  &  2.63 &  0.79 &  0.54 & LRR~\cite{DBLP:journals/pami/LiXWLK23}& 7.12  & 39.16  & 11.41  & 4.37  & 1.46  & 0.45  \\
				MURF~\cite{DBLP:journals/pami/XuYM23} & 5.04 &  16.37 & 8.31 & 2.67  & 0.86  & 0.40  & 	MURF~\cite{DBLP:journals/pami/XuYM23} & 6.91 & 33.34  & 13.88 & 5.37  & 1.04  & 0.52 \\
   			DDFM~\cite{Zhao_2023_ICCV} & 6.19 &  29.26 & 7.44  & 2.51  & 1.45  & 0.73  & DDFM~\cite{Zhao_2023_ICCV}&  7.24 & 42.43  & 10.68  & 4.15  & 1.64  & 0.62  \\
      			SegM~\cite{Liu_2023_ICCV} & 5.95 &  37.28 & 11.10 & 3.47  & 1.57  & 0.88  & SegM~\cite{Liu_2023_ICCV}&  7.29 & 46.14  & 14.47  & 5.57 & 1.61  & 0.65  \\
                    Ours & \textbf{6.71} &  \textbf{44.13} & \textbf{11.56} & \textbf{3.76}  & \textbf{1.63}  & \underline{0.97}  & Ours &  \textbf{7.52} & \textbf{54.81}  & \textbf{15.21}  & \textbf{5.83} & \textbf{1.69}  & \textbf{0.66} \\
			 \midrule
    				\multicolumn{7}{c}{\textbf{Infrared-Visible Image Fusion on M$^3$FD Dataset}~\cite{DBLP:conf/cvpr/LiuFHWLZL22}}                              &                              \multicolumn{7}{c}{\textbf{Medical Image Fusion on Harvard Dataset}~\cite{HarvardMIF}}                               \\
			&  EN $\uparrow$   &   SD $\uparrow$   &  SF $\uparrow$   &  AG $\uparrow$  & SCD $\uparrow$  & VIF  $\uparrow$  &                                        &  EN $\uparrow$   &   SD $\uparrow$   &  SF $\uparrow$   &  AG $\uparrow$  & SCD $\uparrow$  & VIF  $\uparrow$  \\ \midrule
			SDN~\cite{DBLP:journals/ijcv/ZhangM21}  & 6.87  & 36.22  & 15.32  & 5.61  & 1.41  & 0.55  & SDN~\cite{DBLP:journals/ijcv/ZhangM21}  & 3.79  & 52.53 & 21.91 & 5.51  & 0.87  & 0.52   \\
			TarD~\cite{DBLP:conf/cvpr/LiuFHWLZL22}  & 6.80  & 41.77   & 8.65   & 3.17   & 1.35   & 0.51  & TarD~\cite{DBLP:conf/cvpr/LiuFHWLZL22} & \underline{4.74}  & 55.73 & 18.02 & 5.35  & 0.86  & 0.31  \\
			 DeF~\cite{Liang2022ECCV}  & 6.90   & 36.81   & 9.85   & 3.65   & \underline{1.42}   & 0.58   &  DeF~\cite{Liang2022ECCV}   & 4.00  & 57.48 & 17.09 & 4.19  & 0.84  & 0.59 \\
			Meta~\cite{DBLP:conf/cvpr/ZhaoXZHL23}  & 6.73   & 30.56   &  16.48  & \underline{6.02}   & 1.31   & 0.65   & Meta~\cite{DBLP:conf/cvpr/ZhaoXZHL23} & 3.90  & 65.18 & \textbf{28.69} & \textbf{6.29} & 1.33  & 0.54 \\
			CDDF~\cite{Zhao_2023_CVPR} & \underline{7.04}  & \underline{42.02}  & \underline{16.56}  & 5.84   & 1.41 & 0.65  & CDDF~\cite{Zhao_2023_CVPR} & 4.13  & \underline{68.46} & 21.58 & 5.83  & \underline{1.61}  & 0.66   \\
			LRR~\cite{DBLP:journals/pami/LiXWLK23} & 6.58   & 30.28   & 11.83   &  4.21  &  1.34  &  0.54  & LRR~\cite{DBLP:journals/pami/LiXWLK23}& 4.15  & 45.71 & 17.39 & 4.47  & 0.23  & 0.51  \\
				MURF~\cite{DBLP:journals/pami/XuYM23} & 6.59  &  28.89  & 11.82  & 4.81   & 1.21   & 0.39   & 	MURF~\cite{DBLP:journals/pami/XuYM23}& 4.42  & 36.35 & \underline{27.18} & 5.98  & 0.35  & 0.37 \\
   			DDFM~\cite{Zhao_2023_ICCV} & 6.82  &  32.68  & 10.07   & 3.71   & 1.35  & 0.60  & DDFM~\cite{Zhao_2023_ICCV}& 3.97  & 59.81 & 16.43 & 4.11  & 1.49  & 0.63   \\
      			SegM~\cite{Liu_2023_ICCV} & 6.88  &  36.20  & 16.19  & 5.83   & 1.38   & \textbf{0.75}   & SegM~\cite{Liu_2023_ICCV}& 3.67  & 57.79 & 21.91 & 5.56  & 1.05  & \underline{0.66}   \\
                    Ours & \textbf{7.12}  &  \textbf{44.01}  & \textbf{16.92}  & \textbf{6.23}   & \textbf{1.48}   & \underline{0.66} &  Ours& \textbf{4.81} & \textbf{69.42} & 22.15 & \underline{6.02}  & \textbf{1.64} & \textbf{0.66}   \\
			 \bottomrule    
		\end{tabular}}
		\caption{Quantitative results of IVF and MIF task. Best and second-best values are \textbf{highlighted} and \underline{underlined}.}
		\label{tab:Quantitative}%
		\vspace{-1.4em}
	\end{table*}%

	\begin{table}[t]
	\centering
	\resizebox{\linewidth}{!}{
		\begin{tabular}{cccccccc}
			\toprule
			\multirow{2}{*}{}            &                                   {Configurations}                                    &     {EN}      &      {SD}      &     {SF}      &     {AG}      &    {SCD}     &     {VIF}     \\ \midrule
			\uppercase\expandafter{\romannumeral1} &                                 w/o  Equivariant Loss                                 &     6.36      &     39.22      &     9.09      &     3.01      &     1.18     &     0.72      \\
			\uppercase\expandafter{\romannumeral2} &                                   w/o  Sensing Loss                                   &     6.42      &     40.12      &     9.12      &     3.29      &     1.24      &     0.79      \\
			\uppercase\expandafter{\romannumeral3} & w/ $\ell_1(\boldsymbol{f},\boldsymbol{i})+\ell_1(\boldsymbol{f},\boldsymbol{v})$ Loss &     6.21      &     38.96      &     8.85      &     2.99      &     0.85      &     0.76      \\
			\uppercase\expandafter{\romannumeral4} &             Exp.~\uppercase\expandafter{\romannumeral3} w/ augmentations              &     6.26      &     39.11      &     8.73      &     3.02      &     0.96      &     0.77      \\
			\uppercase\expandafter{\romannumeral5} &                                 w/o Global Extractor                                  &     6.45      &     39.37      &     9.44      &     3.24      &     1.42      &     0.81      \\
			\uppercase\expandafter{\romannumeral6} &                                 w/o Local  Extractor                                  &     6.52      &     40.49      &     9.79      &     2.82      &     1.46      &     0.82      \\ \midrule
			&                                         Ours                                          & \textbf{6.71} & \textbf{44.13} & \textbf{11.56} & \textbf{3.76} & \textbf{1.63} & \textbf{0.97} \\ \bottomrule
	\end{tabular}}
	\caption{Ablation experiment results. \textbf{Bold} indicates the best value.}
	\label{tab:ablation}%
	\vspace{-1.5em}
\end{table}

	\section{Experiment}
	\subsection{Infrared and visible image fusion}
	\bfsection{Setup}
	We conduct experiments on three fashion benchmarks: MSRS~\cite{DBLP:journals/inffus/TangYZJM22}, RoadScene~\cite{xu2020aaai} and M$^3$FD~\cite{DBLP:conf/cvpr/LiuFHWLZL22}. The network is trained on the MSRS training set and tested on its test set to evaluate the performance. In addition, the trained model is implemented to RoadScene and M$^3$FD without fine-tuning to verify the generalization performance. Our experiments are performed using PyTorch on a computer equipped with two NVIDIA GeForce RTX 3090 GPUs. The training image pairs are cropped into 128$\times$128 patches randomly and with a batchsize of 8 before being fed into the network. {$\alpha_1$ and $\alpha_2$ in \cref{eq:total_loss} are set to 1 and 0.1}, to ensure comparable magnitudes among the terms in the loss function. We train the network for 100 epochs using the Adam optimizer, with an initial learning rate of 1e-4 and decreasing by a factor of 0.5 every 20 epochs. U-Fuser is set to contain a four-layer structure. $\mathcal{A}_i$ and $\mathcal{A}_v$ are set as five-layer U-Nets~\cite{DBLP:conf/miccai/RonnebergerFB15}. They are pre-trained and parameter-frozen prior to the U-Fuser training. As for the transformation set $\mathcal{G}$, we will discuss it in our supplementary material.

	\begin{figure}[t]
		\centering
		\includegraphics[width=\linewidth]{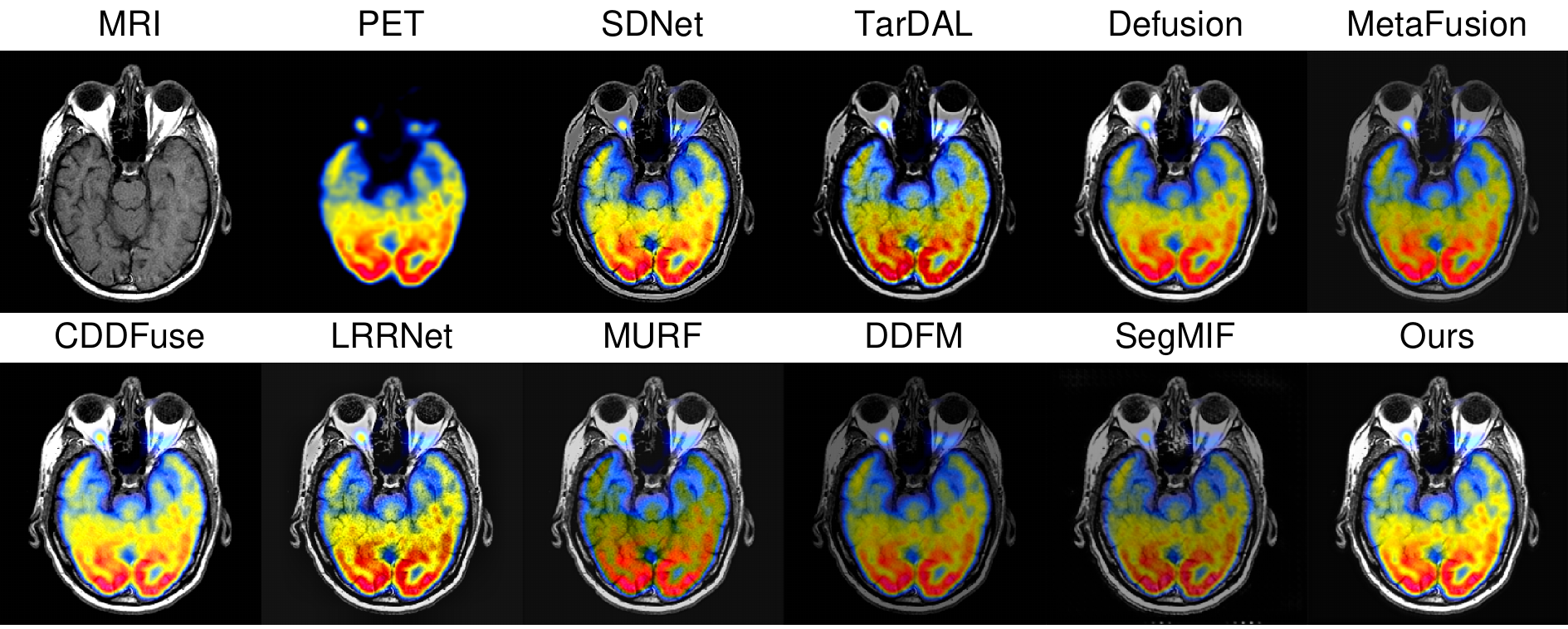}
		\caption{Visual comparison for MIF task.}
		\label{fig:MIF1}
		\vspace{-2em}
	\end{figure}

        \bfsection{SOTA methods and metrics}
	We compare EMMA with SOTA methods of IVF, including 
	SDNet~\cite{DBLP:journals/ijcv/ZhangM21},
	TarDAL~\cite{DBLP:conf/cvpr/LiuFHWLZL22},
	DeFusion~\cite{Liang2022ECCV},
	MetaFusion~\cite{DBLP:conf/cvpr/ZhaoXZHL23},
	CDDFuse~\cite{Zhao_2023_CVPR},
	LRRNet~\cite{DBLP:journals/pami/LiXWLK23},
	MURF~\cite{DBLP:journals/pami/XuYM23},
	DDFM~\cite{Zhao_2023_ICCV} and
	SegMIF~\cite{Liu_2023_ICCV}.
	Six metrics are used to objectively compare fusion performance, including entropy (EN), standard deviation (SD), spatial frequency (SF), average gradient (AG), structure content dissimilarity (SCD) and visual information fidelity (VIF). Higher values indicate superior fusion effects and the calculation details are in \cite{ma2019infrared}.
	
	\bfsection{Qualitative comparison}
	We compare the fusion outcomes of EMMA with SOTAs in \cref{fig:IVF1,fig:IVF2}. Our results successfully integrate thermal radiation information derived from infrared images with detailed texture features extracted from visible images. \cref{fig:IVF1} 
	shows
	that the fused image accurately captures the advantages of each modality while eliminating redundant information. The fusion process enhances object visibility, sharpens textures, and reduces artifacts. In \cref{fig:IVF2}, objects situated in inadequately illuminated surroundings are prominently highlighted with well-defined edges and abundant contours. This distinctiveness facilitates the differentiation between foreground objects and the background, thereby enhancing our comprehension of the depicted scene.
	
	\bfsection{Quantitative comparison}
	The fusion outcomes are quantitatively compared using six metrics, as shown in \cref{tab:Quantitative}. Our method exhibits remarkable performance across nearly all metrics, affirming its suitability for various environmental conditions and object categories. They indicate the capability of EMMA to produce images that align with human visual perception while preserving the integrity of the source image features and producing informative fused images.

	\begin{table*}[t]
	\begin{minipage}[t]{0.445\textwidth}
		\centering
		\resizebox{\linewidth}{!}{
			\begin{tabular}{cccccccc}
				\toprule
				&       Bus       &       Car       &       Lam       &       Mot       &       Peo       &       Tru       &     mAP@0.5     \\ \midrule
				IR    & 78.8  & 88.7  & 70.2  & 63.4  & 80.9  & 65.8  & 74.6  \\
				VI    & 78.3  & 90.7  & 86.4  & 69.3  & 70.5  & 70.9  & 77.7  \\
				SDN~\cite{DBLP:journals/ijcv/ZhangM21} & 81.4  & 92.3  & 84.1  & 67.4  & 79.4  & 69.3  & 79.0  \\
				TarD~\cite{DBLP:conf/cvpr/LiuFHWLZL22}& 81.3  & \textbf{94.8} & 87.1  & 69.3  & 81.5 & 68.7  & 80.5  \\
				DeF~\cite{Liang2022ECCV}&
				82.9 & 92.5  & \textbf{87.8} & 69.5  & 80.8  &
				71.4&
				80.8 \\
				Meta~\cite{DBLP:conf/cvpr/ZhaoXZHL23} & \underline{83.0}  & 93.4  & 87.3  & \underline{74.8}  & 81.6 & 68.8  & 81.5  \\
				CDDF~\cite{Zhao_2023_CVPR} & 81.8  & 92.9  & 87.6  & 72.8  & \underline{81.8}  & \underline{72.9}  & \underline{81.6}  \\		
				LRR~\cite{DBLP:journals/pami/LiXWLK23}  & 80.1  & 92.3  & 86.2  &
				73.6 & 78.3  & 68.6  & 79.9  \\
				MURF~\cite{DBLP:journals/pami/XuYM23} & 81.3  & 92.6  & 86.5  & 70.8  & 80.2  & 69.9  & 80.2  \\
				DDFM~\cite{Zhao_2023_ICCV} & 82.2  & 93.2  & 87.6  & 68.4  & 81.0  & 71.3  & 80.6 \\			
				SegM~\cite{Liu_2023_ICCV} & 81.8  & 93.1  & 86.8  & 72.3  & 79.9  & 70.9  & 80.8  \\
				Ours & \textbf{83.2} &
				\underline{93.5} &
				\underline{87.7} & \textbf{77.7} & \textbf{82.0} & \textbf{73.5} & \textbf{82.9} \\				
				\bottomrule
		\end{tabular}}
		\caption{AP@0.5(\%) for MM detection.}
		\label{tab:MMOD}
	\end{minipage}
	\hfill
	\begin{minipage}[t]{0.55\textwidth}%
		\centering
		\resizebox{\linewidth}{!}{
			\begin{tabular}{ccccccccccc}
				\toprule
				&        Unl        &        Car        &        Per        &        Bik        &        Cur        &        CS         &        GD         &        CC         &        Bu         &       mIOU        \\ \midrule
				IR    & 90.5  & 75.6  & 45.4  & 59.4  & 37.2  & 51.0  & 46.4  & 43.5  & 50.2  & 55.4  \\
				VI    & 84.7  & 67.8  & 56.4  & 51.8  & 34.6  & 39.3  & 42.2  & 40.2  & 48.4  & 51.7  \\
				SDN~\cite{DBLP:journals/ijcv/ZhangM21} & 97.3  & 78.4  & 62.5  & 61.7  & 35.7  & 49.3  & 52.4  & 42.2  & 52.9  & 59.2  \\
				TarD~\cite{DBLP:conf/cvpr/LiuFHWLZL22}& 97.1  & 79.1  & 55.4  & 59.0  & 33.6  & 49.4  & 54.9  & 42.6  & 53.5  & 58.3  \\
				DeF~\cite{Liang2022ECCV}& 97.5  & 82.6 & 61.1  & 62.6 & 40.4  & 51.5  & 48.1  & \underline{47.9} & 54.8  & 60.7  \\
				Meta~\cite{DBLP:conf/cvpr/ZhaoXZHL23} & 97.3 & 81.6  & 61.2  & 62.1  & 37.2  & \underline{52.9} & 59.8 & 46.2  & \underline{56.2}  & 61.6  \\
				CDDF~\cite{Zhao_2023_CVPR} & \textbf{97.8} & 82.5  & 63.2 & 62.2  & 40.8  & 52.7  & 56.2  & 45.3  & 58.7 & 62.2  \\
				LRR~\cite{DBLP:journals/pami/LiXWLK23} & 97.4  & 81.2  & 62.4  & 61.9  & 40.3  & 50.7  & 48.1  & 45.3  & 47.3  & 59.4  \\
				MURF~\cite{DBLP:journals/pami/XuYM23}& 97.2  & 81.4  & 62.0  & 60.9  & 39.7 & 52.3  & 55.5  & 46.8  & 56.1  & 61.3  \\
				DDFM~\cite{Zhao_2023_ICCV}& 97.4  & 82.5  & 60.4  & 62.0  & \underline{41.7} & 52.9  & 56.2  & 46.3  & 53.7 & 61.2  \\
				SegM~\cite{Liu_2023_ICCV}& 97.6  & \textbf{84.6}  & \underline{64.8}  &\textbf{63.6} & 40.2 & 52.9  & \underline{59.9}  & 49.4  & 56.2  & \underline{63.2}  \\
				Ours & \underline{97.6} & \underline{84.0} & \textbf{65.2} & \underline{63.1} & \textbf{42.4} & \textbf{53.6} & \bf{60.2} & \textbf{50.5} & \textbf{56.3} & \textbf{63.7}  \\
				\bottomrule
		\end{tabular}}
		\caption{IoU(\%) for MM segmentation.}
		\label{tab:MMSS}
	\end{minipage}
\vspace{-1.5em}
\end{table*}

	\subsubsection{Ablation studies}
 We conduct ablation studies on the MSRS testset to prove the rationality of EMMA, with the results shown in \cref{tab:ablation}.
 
	\bfsection{Terms in loss function}
	In Exp.~\uppercase\expandafter{\romannumeral1}, we eliminate the last term in \cref{eq:total_loss}, which is the equivariant term. Even though the fusion module is capable of completing image fusion, it is unable to constrain the solution space through the equivariant prior. Thus, the network yields weaker results.
	In Exp.~\uppercase\expandafter{\romannumeral2}, we modified the first two terms of \cref{eq:total_loss} to be $\ell_1(\boldsymbol{f},\boldsymbol{i})+\ell_1(\boldsymbol{f},\boldsymbol{v})$, which is the traditional loss in other fusion tasks.
	The first two terms of \cref{eq:total_loss}, \emph{i.e.}, sensing loss, guarantee that the fused image needs to inherit enough information from source images, so that the output pseudo-perceptual imaging result can be closer to the source images.
	While the traditional loss function purely forces the fused image to closely resemble the source images. Results in Exp.~\uppercase\expandafter{\romannumeral2} demonstrate the necessity of sensing loss term.
	In Exp.~\uppercase\expandafter{\romannumeral3}, we replace the loss in \cref{eq:total_loss} with that in Exp.~\uppercase\expandafter{\romannumeral2}. The results indicate that without equivariant loss and sensing loss, relying on $\ell_1(\boldsymbol{f},\boldsymbol{i})+\ell_1(\boldsymbol{f},\boldsymbol{v})$ loss makes it difficult to achieve an ideal fusion network.
	In Exp.~\uppercase\expandafter{\romannumeral4}, to further demonstrate our claim, we employ the same transformation as EMMA for conducting data augmentation (DA) on input images $\boldsymbol{i}$ and $\boldsymbol{v}$, expanding upon the ablation experiment Exp.~\uppercase\expandafter{\romannumeral3}. That is, we employ the same transformation group as EMMA on the original network input, and the fusion training framework follows traditional approaches. Specifically, the loss function becomes:
	$\|\boldsymbol{f}-\boldsymbol{i}\|+\|\boldsymbol{f}-\boldsymbol{v}\|+\|\boldsymbol{f}_t-\boldsymbol{i}_t\|+\|\boldsymbol{f}_t-\boldsymbol{v}_t\|$, where $\boldsymbol{f}_t=T_g\boldsymbol{f}$. Experimental results demonstrate that under the same transformation, there is only a slight improvement for DA on $\boldsymbol{i}$ and $\boldsymbol{v}$. Conversely, in comparison to EMMA, substantial differences in effectiveness are observed. Thus, our equivariant fusion module fundamentally differs from traditional DA, as DA cannot provide additional information gains when learning to image without ground truth.
	
	\bfsection{U-Fuser}
	Then, in Exp.~\uppercase\expandafter{\romannumeral5} and Exp.~\uppercase\expandafter{\romannumeral6}, we separately eliminated the Restormer-block or the Res-block, ensuring a consistent number of {parameters} by increasing the remaining blocks number. The results demonstrate that an incomplete feature extraction module leads to deficiencies in modeling local texture details or capturing long-range dependencies, thereby resulting in a degradation of performance.

	\subsection{Downstream IVF applications}
	This section aims to examine the impact of image fusion on downstream vision tasks. We assess the performance of fusion results in both \textit{multi-modal semantic segmentation} (MMSS) tasks and \textit{multi-modal object detection} (MMOD) tasks. To ensure fairness, we individually re-train the network for each task using fusion results obtained from their own methods. Due to space limitations, the visual comparisons are placed in the supplementary material.
 
		\bfsection{Infrared-visible object detection}
	MMOD task is conducted on the M$^3$FD dataset~\cite{DBLP:conf/cvpr/LiuFHWLZL22}, which comprises 4200 images encompassing six categories of labels: people, cars, buses, motorcycles, trucks, and lamps. We partition M$^3$FD dataset into training/validation/test sets in an 8:1:1 ratio. YOLOv5 detector~\cite{glenn_jocher_2020_4154370} is trained using the SGD optimizer for 400 epochs. 
	Batch size is $8$ and the initial learning is $0.01$.
	We evaluate the detection performance by comparing the mAP@0.5.
	\cref{tab:MMOD} indicates that EMMA exhibits the most superior detection capabilities, enhancing the detection accuracy by merging thermal radiation and RGB information and emphasizing hard-to-detect objects.
 
	\bfsection{Infrared-visible semantic segmentation}
	MSRS dataset \cite{DBLP:journals/inffus/TangYZJM22} is designed for MMSS task and encompasses nine categories of pixel-level labels: background, bump, color cone, guardrail, curve, bike, person, car stop, and car. We select DeeplabV3+~\cite{DBLP:conf/eccv/ChenZPSA18} as the segmentation network and value the performances via Intersection over Union (IoU). The division of training and test sets adheres to the protocol in the original dataset paper~\cite{DBLP:journals/inffus/TangYZJM22}. We employ the cross-entropy loss along with the SGD optimizer. The total number of epochs is 340 while the backbone is frozen for the first 100 epochs. The batch size and the initial learning rate are set to 8 and 7e-3, and the learning rate follows cosine annealing delayed as the epoch number increases.
	Segmentation outcomes are displayed in~\cref{tab:MMSS}.
	EMMA effectively combines the edge and contour details present in the source images, thereby improving the model's capability to recognize the object's boundary, and leading to more precise segmentation.

	\subsection{Medical image fusion}
	\bfsection{Setup}
	We conducted MIF experiments via the Harvard Medical dataset \cite{HarvardMIF}, which included 50 pairs of MRI-CT/MRI-PET/MRI-SPECT images. We directly generalize the models trained on the IVF task to the MIF task without fine-tuning. The quantitative metrics used are the same as those employed in the IVF task.
	
	\bfsection{Comparison with SOTA methods}
	In both visual perception and quantitative measures in \cref{fig:MIF1,tab:Quantitative}, EMMA demonstrates superior accuracy in extracting structural highlights and detailed texture features, and effectively integrates characteristic features within the fused image. Consequently, it achieves remarkable fusion results.

	\section{Conclusion}
	This paper tackles the lack of ground truth in image fusion by employing a conceptually straightforward yet potent prior that natural imaging responses exhibit equivariance to translations like shifts, rotations, and reflections. Upon this foundation, we propose a self-supervised paradigm called equivariant image fusion, which adjusts the inherent patterns of the loss function by taking into account the principles of natural imaging, making it simulate the sensing-imaging process. We also introduce a U-Net-like fusion module using the Restormer-CNN block as its basic unit, facilitating global-local feature extraction and efficient information fusion. Experimental results corroborate the effectiveness of our proposed paradigm in multi-modality image fusion, and its propensity to facilitate downstream tasks like multi-modality segmentation and detection.

    \section*{Acknowledgement}\label{sec:6}
    This work has been supported by the National Natural Science Foundation of China under Grant 12371512 and 12201497, 
    the Guangdong Basic and Applied Basic Research Foundation under Grant 2023A1515011358, and partly supported by the Alexander von Humboldt Foundation.	
    {
        \small
        \bibliographystyle{ieeenat_fullname}
        \bibliography{xbib}

\begin{thebibliography}{56}
\providecommand{\natexlab}[1]{#1}
\providecommand{\url}[1]{\texttt{#1}}
\expandafter\ifx\csname urlstyle\endcsname\relax
  \providecommand{\doi}[1]{doi: #1}\else
  \providecommand{\doi}{doi: \begingroup \urlstyle{rm}\Url}\fi

\bibitem[Bochkovskiy et~al.(2020)Bochkovskiy, Wang, and Liao]{DBLP:journals/corr/abs-2004-10934}
Alexey Bochkovskiy, Chien{-}Yao Wang, and Hong{-}Yuan~Mark Liao.
\newblock Yolov4: Optimal speed and accuracy of object detection.
\newblock \emph{CoRR}, abs/2004.10934, 2020.

\bibitem[Chen et~al.(2021)Chen, Tachella, and Davies]{DBLP:conf/iccv/0004TD21}
Dongdong Chen, Juli{\'{a}}n Tachella, and Mike~E. Davies.
\newblock Equivariant imaging: Learning beyond the range space.
\newblock In \emph{{ICCV}}, pages 4359--4368. {IEEE}, 2021.

\bibitem[Chen et~al.(2022)Chen, Tachella, and Davies]{DBLP:conf/cvpr/0004TD22}
Dongdong Chen, Juli{\'{a}}n Tachella, and Mike~E. Davies.
\newblock Robust equivariant imaging: a fully unsupervised framework for learning to image from noisy and partial measurements.
\newblock In \emph{{CVPR}}, pages 5637--5646. {IEEE}, 2022.

\bibitem[Chen et~al.(2023)Chen, Davies, Ehrhardt, Sch{\"{o}}nlieb, Sherry, and Tachella]{DBLP:journals/spm/ChenDESST23}
Dongdong Chen, Mike~E. Davies, Matthias~J. Ehrhardt, Carola{-}Bibiane Sch{\"{o}}nlieb, Ferdia Sherry, and Juli{\'{a}}n Tachella.
\newblock Imaging with equivariant deep learning: From unrolled network design to fully unsupervised learning.
\newblock \emph{{IEEE} Signal Process. Mag.}, 40\penalty0 (1):\penalty0 134--147, 2023.

\bibitem[Chen et~al.(2018)Chen, Zhu, Papandreou, Schroff, and Adam]{DBLP:conf/eccv/ChenZPSA18}
Liang{-}Chieh Chen, Yukun Zhu, George Papandreou, Florian Schroff, and Hartwig Adam.
\newblock Encoder-decoder with atrous separable convolution for semantic image segmentation.
\newblock In \emph{{ECCV}}, pages 833--851. Springer, 2018.

\bibitem[Deng and Dragotti(2021)]{DBLP:journals/pami/0002D21}
Xin Deng and Pier~Luigi Dragotti.
\newblock Deep convolutional neural network for multi-modal image restoration and fusion.
\newblock \emph{{IEEE} Trans. Pattern Anal. Mach. Intell.}, 43\penalty0 (10):\penalty0 3333--3348, 2021.

\bibitem[Fang et~al.(2024)Fang, Wang, and Ye]{fang2024glgnet}
Li Fang, Qian Wang, and Long Ye.
\newblock Glgnet: light field angular superresolution with arbitrary interpolation rates.
\newblock \emph{Visual Intelligence}, 2\penalty0 (1):\penalty0 6, 2024.

\bibitem[Gao et~al.(2022)Gao, Deng, Xu, Xu, and Dragotti]{DBLP:journals/tip/GaoDXXD22}
Fangyuan Gao, Xin Deng, Mai Xu, Jingyi Xu, and Pier~Luigi Dragotti.
\newblock Multi-modal convolutional dictionary learning.
\newblock \emph{{IEEE} Trans. Image Process.}, 31:\penalty0 1325--1339, 2022.

\bibitem[{Harvard Medical website}()]{HarvardMIF}
{Harvard Medical website}.
\newblock \url{http://www.med.harvard.edu/AANLIB/home.html}.

\bibitem[He et~al.(2016)He, Zhang, Ren, and Sun]{HeZRS16}
Kaiming He, Xiangyu Zhang, Shaoqing Ren, and Jian Sun.
\newblock Deep residual learning for image recognition.
\newblock In \emph{Conference on Computer Vision and Pattern Recognition, {CVPR}}, pages 770--778, 2016.

\bibitem[Huang et~al.(2022)Huang, Liu, Fan, Liu, Zhong, and Luo]{huangreconet}
Zhanbo Huang, Jinyuan Liu, Xin Fan, Risheng Liu, Wei Zhong, and Zhongxuan Luo.
\newblock Reconet: Recurrent correction network for fast and efficient multi-modality image fusion.
\newblock In \emph{European Conference on Computer Vision (ECCV)}, 2022.

\bibitem[James and Dasarathy(2014)]{DBLP:journals/inffus/JamesD14}
Alex~Pappachen James and Belur~V. Dasarathy.
\newblock Medical image fusion: {A} survey of the state of the art.
\newblock \emph{Inf. Fusion}, 19:\penalty0 4--19, 2014.

\bibitem[Jiang et~al.(2022)Jiang, Zhang, Fan, and Liu]{DBLP:conf/mm/JiangZ0L22}
Zhiying Jiang, Zengxi Zhang, Xin Fan, and Risheng Liu.
\newblock Towards all weather and unobstructed multi-spectral image stitching: Algorithm and benchmark.
\newblock In \emph{{ACM} Multimedia}, pages 3783--3791, 2022.

\bibitem[Jocher(2020)]{glenn_jocher_2020_4154370}
Glenn Jocher.
\newblock {ultralytics/yolov5}.
\newblock \url{https://github.com/ultralytics/yolov5}, 2020.

\bibitem[Jung et~al.(2020)Jung, Kim, Jang, Ha, and Sohn]{DBLP:journals/tip/JungKJHS20}
Hyungjoo Jung, Youngjung Kim, Hyunsung Jang, Namkoo Ha, and Kwanghoon Sohn.
\newblock Unsupervised deep image fusion with structure tensor representations.
\newblock \emph{{IEEE} Trans. Image Process.}, 29:\penalty0 3845--3858, 2020.

\bibitem[Li and Wu(2018)]{li2018densefuse}
Hui Li and Xiao-Jun Wu.
\newblock Densefuse: A fusion approach to infrared and visible images.
\newblock \emph{IEEE Transactions on Image Processing}, 28\penalty0 (5):\penalty0 2614--2623, 2018.

\bibitem[Li et~al.(2020)Li, Wu, and Kittler]{DBLP:journals/tip/LiWK20}
Hui Li, Xiao{-}Jun Wu, and Josef Kittler.
\newblock Mdlatlrr: {A} novel decomposition method for infrared and visible image fusion.
\newblock \emph{{IEEE} Trans. Image Process.}, 29:\penalty0 4733--4746, 2020.

\bibitem[Li et~al.(2021)Li, Wu, and Kittler]{DBLP:journals/inffus/LiWK21}
Hui Li, Xiao{-}Jun Wu, and Josef Kittler.
\newblock Rfn-nest: An end-to-end residual fusion network for infrared and visible images.
\newblock \emph{Inf. Fusion}, 73:\penalty0 72--86, 2021.

\bibitem[Li et~al.(2023)Li, Xu, Wu, Lu, and Kittler]{DBLP:journals/pami/LiXWLK23}
Hui Li, Tianyang Xu, Xiaojun Wu, Jiwen Lu, and Josef Kittler.
\newblock Lrrnet: {A} novel representation learning guided fusion network for infrared and visible images.
\newblock \emph{{IEEE} Trans. Pattern Anal. Mach. Intell.}, 45\penalty0 (9):\penalty0 11040--11052, 2023.

\bibitem[Liang et~al.(2022)Liang, Jiang, Liu, and Ma]{Liang2022ECCV}
Pengwei Liang, Junjun Jiang, Xianming Liu, and Jiayi Ma.
\newblock Fusion from decomposition: A self-supervised decomposition approach for image fusion.
\newblock In \emph{European Conference on Computer Vision (ECCV)}, 2022.

\bibitem[Liu et~al.(2022)Liu, Fan, Huang, Wu, Liu, Zhong, and Luo]{DBLP:conf/cvpr/LiuFHWLZL22}
Jinyuan Liu, Xin Fan, Zhanbo Huang, Guanyao Wu, Risheng Liu, Wei Zhong, and Zhongxuan Luo.
\newblock Target-aware dual adversarial learning and a multi-scenario multi-modality benchmark to fuse infrared and visible for object detection.
\newblock In \emph{{CVPR}}, pages 5792--5801. {IEEE}, 2022.

\bibitem[Liu et~al.(2023{\natexlab{a}})Liu, Lin, Wu, Liu, Luo, and Fan]{liu2023coconet}
Jinyuan Liu, Runjia Lin, Guanyao Wu, Risheng Liu, Zhongxuan Luo, and Xin Fan.
\newblock Coconet: Coupled contrastive learning network with multi-level feature ensemble for multi-modality image fusion.
\newblock \emph{International Journal of Computer Vision}, pages 1--28, 2023{\natexlab{a}}.

\bibitem[Liu et~al.(2023{\natexlab{b}})Liu, Liu, Wu, Ma, Liu, Zhong, Luo, and Fan]{Liu_2023_ICCV}
Jinyuan Liu, Zhu Liu, Guanyao Wu, Long Ma, Risheng Liu, Wei Zhong, Zhongxuan Luo, and Xin Fan.
\newblock Multi-interactive feature learning and a full-time multi-modality benchmark for image fusion and segmentation.
\newblock In \emph{Proceedings of the IEEE/CVF International Conference on Computer Vision (ICCV)}, pages 8115--8124, 2023{\natexlab{b}}.

\bibitem[Liu et~al.(2021)Liu, Liu, Liu, and Fan]{DBLP:conf/mm/LiuLL021}
Risheng Liu, Zhu Liu, Jinyuan Liu, and Xin Fan.
\newblock Searching a hierarchically aggregated fusion architecture for fast multi-modality image fusion.
\newblock In \emph{{ACM} Multimedia}, pages 1600--1608. {ACM}, 2021.

\bibitem[Ma et~al.(2019{\natexlab{a}})Ma, Ma, and Li]{ma2019infrared}
Jiayi Ma, Yong Ma, and Chang Li.
\newblock Infrared and visible image fusion methods and applications: A survey.
\newblock \emph{Information Fusion}, 45:\penalty0 153--178, 2019{\natexlab{a}}.

\bibitem[Ma et~al.(2019{\natexlab{b}})Ma, Yu, Liang, Li, and Jiang]{ma2019fusiongan}
Jiayi Ma, Wei Yu, Pengwei Liang, Chang Li, and Junjun Jiang.
\newblock Fusiongan: A generative adversarial network for infrared and visible image fusion.
\newblock \emph{Information Fusion}, 48:\penalty0 11--26, 2019{\natexlab{b}}.

\bibitem[Ma et~al.(2020{\natexlab{a}})Ma, Liang, Yu, Chen, Guo, Wu, and Jiang]{ma2020infrared}
Jiayi Ma, Pengwei Liang, Wei Yu, Chen Chen, Xiaojie Guo, Jia Wu, and Junjun Jiang.
\newblock Infrared and visible image fusion via detail preserving adversarial learning.
\newblock \emph{Information Fusion}, 54:\penalty0 85--98, 2020{\natexlab{a}}.

\bibitem[Ma et~al.(2020{\natexlab{b}})Ma, Xu, Jiang, Mei, and Zhang]{DBLP:journals/tip/MaXJMZ20}
Jiayi Ma, Han Xu, Junjun Jiang, Xiaoguang Mei, and Xiao{-}Ping~(Steven) Zhang.
\newblock Ddcgan: {A} dual-discriminator conditional generative adversarial network for multi-resolution image fusion.
\newblock \emph{{IEEE} Trans. Image Process.}, 29:\penalty0 4980--4995, 2020{\natexlab{b}}.

\bibitem[Meher et~al.(2019)Meher, Agrawal, Panda, and Abraham]{meher2019a}
Bikash Meher, Sanjay Agrawal, Rutuparna Panda, and Ajith Abraham.
\newblock A survey on region based image fusion methods.
\newblock \emph{Information Fusion}, 48:\penalty0 119--132, 2019.

\bibitem[Ronneberger et~al.(2015)Ronneberger, Fischer, and Brox]{DBLP:conf/miccai/RonnebergerFB15}
Olaf Ronneberger, Philipp Fischer, and Thomas Brox.
\newblock U-net: Convolutional networks for biomedical image segmentation.
\newblock In \emph{{MICCAI}}, pages 234--241. Springer, 2015.

\bibitem[Sun et~al.(2022)Sun, Cao, Zhu, and Hu]{DBLP:conf/mm/SunCZH22}
Yiming Sun, Bing Cao, Pengfei Zhu, and Qinghua Hu.
\newblock Detfusion: {A} detection-driven infrared and visible image fusion network.
\newblock In \emph{{ACM} Multimedia}, pages 4003--4011, 2022.

\bibitem[Tachella et~al.(2023)Tachella, Chen, and Davies]{JMLR:v24:22-0315}
Juli{\'{a}}n Tachella, Dongdong Chen, and Mike Davies.
\newblock Sensing theorems for unsupervised learning in linear inverse problems.
\newblock \emph{Journal of Machine Learning Research}, 24\penalty0 (39):\penalty0 1--45, 2023.

\bibitem[Tang et~al.(2022{\natexlab{a}})Tang, Yuan, and Ma]{DBLP:journals/inffus/TangYM22}
Linfeng Tang, Jiteng Yuan, and Jiayi Ma.
\newblock Image fusion in the loop of high-level vision tasks: {A} semantic-aware real-time infrared and visible image fusion network.
\newblock \emph{Inf. Fusion}, 82:\penalty0 28--42, 2022{\natexlab{a}}.

\bibitem[Tang et~al.(2022{\natexlab{b}})Tang, Yuan, Zhang, Jiang, and Ma]{DBLP:journals/inffus/TangYZJM22}
Linfeng Tang, Jiteng Yuan, Hao Zhang, Xingyu Jiang, and Jiayi Ma.
\newblock Piafusion: {A} progressive infrared and visible image fusion network based on illumination aware.
\newblock \emph{Inf. Fusion}, 83-84:\penalty0 79--92, 2022{\natexlab{b}}.

\bibitem[Vs et~al.(2022)Vs, Valanarasu, Oza, and Patel]{vs2022image}
Vibashan Vs, Jeya Maria~Jose Valanarasu, Poojan Oza, and Vishal~M Patel.
\newblock Image fusion transformer.
\newblock In \emph{2022 IEEE International Conference on Image Processing (ICIP)}, pages 3566--3570. IEEE, 2022.

\bibitem[Wang et~al.(2022)Wang, Liu, Fan, and Liu]{DBLP:conf/ijcai/WangLFL22}
Di Wang, Jinyuan Liu, Xin Fan, and Risheng Liu.
\newblock Unsupervised misaligned infrared and visible image fusion via cross-modality image generation and registration.
\newblock In \emph{{IJCAI}}, pages 3508--3515. ijcai.org, 2022.

\bibitem[Xu et~al.(2020{\natexlab{a}})Xu, Ma, Le, Jiang, and Guo]{xu2020aaai}
Han Xu, Jiayi Ma, Zhuliang Le, Junjun Jiang, and Xiaojie Guo.
\newblock Fusiondn: A unified densely connected network for image fusion.
\newblock In \emph{{AAAI} Conference on Artificial Intelligence, {AAAI}}, pages 12484--12491, 2020{\natexlab{a}}.

\bibitem[Xu et~al.(2022{\natexlab{a}})Xu, Ma, Jiang, Guo, and Ling]{9151265}
Han Xu, Jiayi Ma, Junjun Jiang, Xiaojie Guo, and Haibin Ling.
\newblock U2fusion: {A} unified unsupervised image fusion network.
\newblock \emph{{IEEE} Trans. Pattern Anal. Mach. Intell.}, 44\penalty0 (1):\penalty0 502--518, 2022{\natexlab{a}}.

\bibitem[Xu et~al.(2022{\natexlab{b}})Xu, Ma, Yuan, Le, and Liu]{DBLP:conf/cvpr/Xu0YLL22}
Han Xu, Jiayi Ma, Jiteng Yuan, Zhuliang Le, and Wei Liu.
\newblock Rfnet: Unsupervised network for mutually reinforcing multi-modal image registration and fusion.
\newblock In \emph{{CVPR}}, pages 19647--19656. {IEEE}, 2022{\natexlab{b}}.

\bibitem[Xu et~al.(2023)Xu, Yuan, and Ma]{DBLP:journals/pami/XuYM23}
Han Xu, Jiteng Yuan, and Jiayi Ma.
\newblock {MURF:} mutually reinforcing multi-modal image registration and fusion.
\newblock \emph{{IEEE} Trans. Pattern Anal. Mach. Intell.}, 45\penalty0 (10):\penalty0 12148--12166, 2023.

\bibitem[Xu et~al.(2020{\natexlab{b}})Xu, Zhao, Wang, Zhang, Liu, and Zhang]{DBLP:journals/corr/abs-2005-08448}
Shuang Xu, Zixiang Zhao, Yicheng Wang, Chunxia Zhang, Junmin Liu, and Jiangshe Zhang.
\newblock Deep convolutional sparse coding networks for image fusion.
\newblock \emph{CoRR}, abs/2005.08448, 2020{\natexlab{b}}.

\bibitem[Yan et~al.(2022{\natexlab{a}})Yan, Wang, Li, Zhang, Li, Li, and Yang]{yan2022learning}
Zhiqiang Yan, Kun Wang, Xiang Li, Zhenyu Zhang, Guangyu Li, Jun Li, and Jian Yang.
\newblock Learning complementary correlations for depth super-resolution with incomplete data in real world.
\newblock \emph{IEEE transactions on neural networks and learning systems}, 2022{\natexlab{a}}.

\bibitem[Yan et~al.(2022{\natexlab{b}})Yan, Wang, Li, Zhang, Li, and Yang]{yan2022rignet}
Zhiqiang Yan, Kun Wang, Xiang Li, Zhenyu Zhang, Jun Li, and Jian Yang.
\newblock Rignet: Repetitive image guided network for depth completion.
\newblock In \emph{European Conference on Computer Vision}, pages 214--230. Springer, 2022{\natexlab{b}}.

\bibitem[Ye et~al.(2023)Ye, Yan, Gao, and Yang]{ye2023lfienet}
Wuyang Ye, Tao Yan, Jiahui Gao, and Yang Yang.
\newblock Lfienet: Light field image enhancement network by fusing exposures of lf-dslr image pairs.
\newblock \emph{IEEE Transactions on Computational Imaging}, 2023.

\bibitem[Zamir et~al.(2022)Zamir, Arora, Khan, Hayat, Khan, and Yang]{DBLP:conf/cvpr/ZamirA0HK022}
Syed~Waqas Zamir, Aditya Arora, Salman Khan, Munawar Hayat, Fahad~Shahbaz Khan, and Ming{-}Hsuan Yang.
\newblock Restormer: Efficient transformer for high-resolution image restoration.
\newblock In \emph{{CVPR}}, pages 5718--5729. {IEEE}, 2022.

\bibitem[Zhang and Ma(2021)]{DBLP:journals/ijcv/ZhangM21}
Hao Zhang and Jiayi Ma.
\newblock Sdnet: {A} versatile squeeze-and-decomposition network for real-time image fusion.
\newblock \emph{Int. J. Comput. Vis.}, 129\penalty0 (10):\penalty0 2761--2785, 2021.

\bibitem[Zhang et~al.(2020{\natexlab{a}})Zhang, Xu, Xiao, Guo, and Ma]{DBLP:conf/aaai/ZhangXXGM20}
Hao Zhang, Han Xu, Yang Xiao, Xiaojie Guo, and Jiayi Ma.
\newblock Rethinking the image fusion: {A} fast unified image fusion network based on proportional maintenance of gradient and intensity.
\newblock In \emph{{AAAI}}, pages 12797--12804. {AAAI} Press, 2020{\natexlab{a}}.

\bibitem[Zhang and Demiris(2023)]{10088423}
Xingchen Zhang and Yiannis Demiris.
\newblock Visible and infrared image fusion using deep learning.
\newblock \emph{IEEE Transactions on Pattern Analysis and Machine Intelligence}, pages 1--20, 2023.

\bibitem[Zhang et~al.(2020{\natexlab{b}})Zhang, Liu, Sun, Yan, Zhao, and Zhang]{DBLP:journals/inffus/ZhangLSYZZ20}
Yu Zhang, Yu Liu, Peng Sun, Han Yan, Xiaolin Zhao, and Li Zhang.
\newblock {IFCNN:} {A} general image fusion framework based on convolutional neural network.
\newblock \emph{Inf. Fusion}, 54:\penalty0 99--118, 2020{\natexlab{b}}.

\bibitem[Zhao et~al.(2023{\natexlab{a}})Zhao, Xie, Zhao, He, and Lu]{DBLP:conf/cvpr/ZhaoXZHL23}
Wenda Zhao, Shigeng Xie, Fan Zhao, You He, and Huchuan Lu.
\newblock Metafusion: Infrared and visible image fusion via meta-feature embedding from object detection.
\newblock In \emph{{CVPR}}, pages 13955--13965. {IEEE}, 2023{\natexlab{a}}.

\bibitem[Zhao et~al.(2020)Zhao, Xu, Zhang, Liu, Zhang, and Li]{zhaoijcai2020}
Zixiang Zhao, Shuang Xu, Chunxia Zhang, Junmin Liu, Jiangshe Zhang, and Pengfei Li.
\newblock {DIDFuse}: Deep image decomposition for infrared and visible image fusion.
\newblock In \emph{International Joint Conference on Artificial Intelligence, {IJCAI}}, pages 970--976, 2020.

\bibitem[Zhao et~al.(2022{\natexlab{a}})Zhao, Xu, Zhang, Liang, Zhang, and Liu]{DBLP:journals/tcsv/ZhaoXZLZL22}
Zixiang Zhao, Shuang Xu, Jiangshe Zhang, Chengyang Liang, Chunxia Zhang, and Junmin Liu.
\newblock Efficient and model-based infrared and visible image fusion via algorithm unrolling.
\newblock \emph{{IEEE} Trans. Circuits Syst. Video Technol.}, 32\penalty0 (3):\penalty0 1186--1196, 2022{\natexlab{a}}.

\bibitem[Zhao et~al.(2022{\natexlab{b}})Zhao, Zhang, Xu, Lin, and Pfister]{DBLP:journals/corr/abs-2104-06977}
Zixiang Zhao, Jiangshe Zhang, Shuang Xu, Zudi Lin, and Hanspeter Pfister.
\newblock Discrete cosine transform network for guided depth map super-resolution.
\newblock In \emph{Proceedings of the IEEE/CVF Conference on Computer Vision and Pattern Recognition (CVPR)}, pages 5697--5707, 2022{\natexlab{b}}.

\bibitem[Zhao et~al.(2023{\natexlab{b}})Zhao, Bai, Zhang, Zhang, Xu, Lin, Timofte, and Van~Gool]{Zhao_2023_CVPR}
Zixiang Zhao, Haowen Bai, Jiangshe Zhang, Yulun Zhang, Shuang Xu, Zudi Lin, Radu Timofte, and Luc Van~Gool.
\newblock Cddfuse: Correlation-driven dual-branch feature decomposition for multi-modality image fusion.
\newblock In \emph{Proceedings of the IEEE/CVF Conference on Computer Vision and Pattern Recognition (CVPR)}, pages 5906--5916, 2023{\natexlab{b}}.

\bibitem[Zhao et~al.(2023{\natexlab{c}})Zhao, Bai, Zhu, Zhang, Xu, Zhang, Zhang, Meng, Timofte, and Van~Gool]{Zhao_2023_ICCV}
Zixiang Zhao, Haowen Bai, Yuanzhi Zhu, Jiangshe Zhang, Shuang Xu, Yulun Zhang, Kai Zhang, Deyu Meng, Radu Timofte, and Luc Van~Gool.
\newblock Ddfm: Denoising diffusion model for multi-modality image fusion.
\newblock In \emph{Proceedings of the IEEE/CVF International Conference on Computer Vision (ICCV)}, pages 8082--8093, 2023{\natexlab{c}}.

\bibitem[Zhao et~al.(2023{\natexlab{d}})Zhao, Zhang, Gu, Tan, Xu, Zhang, Timofte, and Van~Gool]{Zhao_2023_ICCV2}
Zixiang Zhao, Jiangshe Zhang, Xiang Gu, Chengli Tan, Shuang Xu, Yulun Zhang, Radu Timofte, and Luc Van~Gool.
\newblock Spherical space feature decomposition for guided depth map super-resolution.
\newblock In \emph{Proceedings of the IEEE/CVF International Conference on Computer Vision (ICCV)}, pages 12547--12558, 2023{\natexlab{d}}.

\end{thebibliography}
    }

\end{document}